\documentclass[]{bytedance_seed}

\makeatletter 
\AtBeginDocument{%
  \setlength{\headheight}{34pt}%
  \setlength{\headsep}{18pt}%
}

\let\old@maketitle\maketitle
\renewcommand{\maketitle}{%
  \old@maketitle
  \vspace{8pt} 
}
\makeatother

\usepackage[utf8]{inputenc}
\usepackage[T1]{fontenc}

\usepackage{amsmath}
\usepackage{amsfonts}
\usepackage{amssymb}
\usepackage{bbm}
\usepackage{nicefrac}

\usepackage{graphicx}
\usepackage{subcaption}
\usepackage{wrapfig}
\graphicspath{{figure/}}

\usepackage{booktabs}
\usepackage{multirow}
\usepackage{threeparttable}
\usepackage{tabularx}
\usepackage{array}
\usepackage{makecell}
\usepackage{longtable}
\usepackage{adjustbox}
\usepackage{arydshln}

\usepackage[table]{xcolor}

\definecolor{bestblue}{RGB}{222,236,255}
\definecolor{gainred}{HTML}{B22222}
\definecolor{lossblue}{HTML}{1F4E8C}
\definecolor{neutralgray}{HTML}{777777}
\definecolor{osured}{RGB}{187,0,0}

\usepackage[ruled]{algorithm2e}
\usepackage{algpseudocode}

\usepackage{microtype}
\usepackage{float}
\usepackage{enumitem}
\usepackage{multicol}

\usepackage{hyperref}
\usepackage{url}
\usepackage{cleveref}

\crefname{section}{sec.}{secs.}
\Crefname{section}{Sec.}{Secs.}
\crefname{figure}{fig.}{figs.}
\Crefname{figure}{Fig.}{Figs.}
\usepackage{fvextra}

\usepackage[most]{tcolorbox}
\tcbuselibrary{listings,breakable,skins}

\definecolor{SeedBlue}{RGB}{187,0,0}
\definecolor{PrimaryColor}{RGB}{187,0,0}
\definecolor{AccentColor}{RGB}{187,0,0}

\newcommand{\harnesscell}[2]{%
  \multirow{#1}{*}{%
    \rotatebox[origin=c]{90}{\makecell[c]{#2}}
  }%
}

\newcommand{\pos}[1]{%
    \textcolor{gainred}{+#1}
}

\newcommand{\down}[1]{%
    \textcolor{lossblue}{#1}
}

\newcommand{\zero}[1]{%
    \textcolor{neutralgray}{#1}
}

\newcommand{\NA}{%
    \textcolor{neutralgray}{--}}
    
\newcommand{\cond}[1]{%
    \textsc{#1}
}

\newcommand{\skillevolbench}{SkillEvolBench}



\newtcolorbox{promptbox}[2][Judge Prompt]{
  colback=black!5!white,
  arc=5pt,
  boxrule=0.6pt,
  fonttitle=\bfseries,
  title=#1,
  before upper={\small},
  colframe=osured,
  label=#2,
}


\title{\skillevolbench:  Benchmarking the Evolution from Episodic Experience to Procedural Skills}

\author{
Yingtie Lei\textsuperscript{1*},
Zhongwei Wan\textsuperscript{1*},
Jiankun Zhang\textsuperscript{2},
Samiul Alam\textsuperscript{1},
Zixuan Zhong\textsuperscript{3},
Peizhou Huang\textsuperscript{4},
Xin Wang\textsuperscript{1},
Jingxuan Zhang\textsuperscript{1},
Donghao Zhou\textsuperscript{5},
Yunta Hsieh\textsuperscript{4},
Zhihao Dou\textsuperscript{6},
Hui Shen\textsuperscript{4},
Yan Xu\textsuperscript{7},
Dimitrios Dimitriadis\textsuperscript{7},
Tuo Zhang\textsuperscript{7},
Mi Zhang\textsuperscript{1}
}

\affiliation{%
  {\fontsize{9}{11}\selectfont
  \textsuperscript{1}The Ohio State University,
  \textsuperscript{2}The University of Chicago,
  \textsuperscript{3}University College London,
  \textsuperscript{4}University of Michigan,
  \textsuperscript{5}The Chinese University of Hong Kong,
  \textsuperscript{6}Case Western Reserve University,
  \textsuperscript{7}Amazon\\[2pt]
  
  \textsuperscript{*} Equal contribution\\[2pt]
  
  {\textbf{Correspondence:} 
  Tuo Zhang \href{mailto:tuozhang@amazon.com}{\texttt{tuozhang@amazon.com}},
  Mi Zhang \href{mailto:mizhang.1@osu.edu}{\texttt{mizhang.1@osu.edu}}}\\
  
  {\textbf{Project Page:} 
  \url{https://skillevolbench.github.io/}}
  }%
}

\abstract{
Large language model (LLM) agents accumulate rich episodic trajectories while solving real-world tasks, but it remains unclear whether such experience can be distilled into reusable procedural skills. We introduce \textbf{\skillevolbench}, a diagnostic benchmark for evaluating this step from experience reuse to skill formation. It contains 180 tasks across six real-world agent environments, organized into role-conditioned task families with shared latent procedures. Agents learn from acquisition tasks, update an external skill library using compacted trajectories and verifier feedback, and then face frozen deployment tasks testing context shift, adversarial shortcuts, and composition. By comparing self-generated and curated-start skill evolution against no-skill and raw-trajectory controls, \skillevolbench~separates procedural abstraction from base capability, curated prior knowledge, and direct reuse of episodic traces. Across ten model configurations and three agent harnesses, we find that current agents often adapt locally but rarely form robust reusable skills. Skill-based conditions can improve acquisition or replay, and individual models sometimes gain on specific deployment axes, but these gains are unstable under frozen deployment. Raw-trajectory reuse frequently outperforms distilled skills, suggesting that current abstraction procedures discard contextual and procedural cues that remain useful for future tasks. Capacity and cost analyses further show that writing more skills or larger Tier-3 resource libraries is not sufficient: additional updates can improve coverage while introducing episode-specific drift and procedural clutter. These findings position \skillevolbench~as a testbed for measuring when one-off experience becomes durable procedural knowledge rather than task-local memory.
}

\begin{document}
\maketitle
\begin{figure*}[!h]
    \centering
    \includegraphics[width=\textwidth]{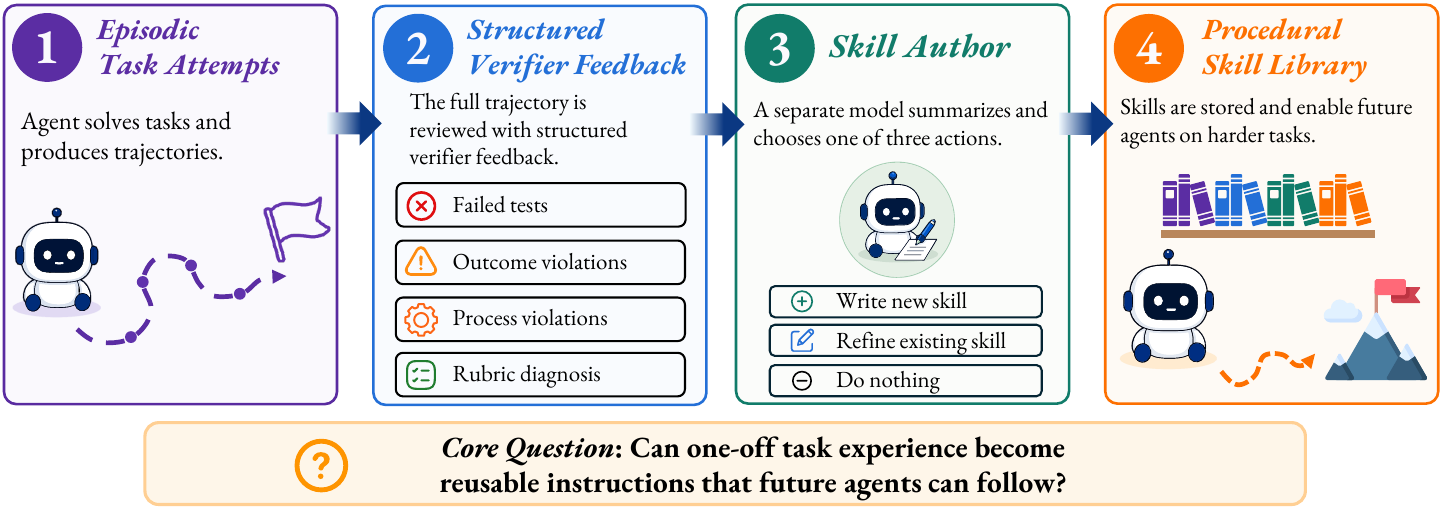}
\caption{
\textbf{\textsc{SkillEvolBench} probes whether episodic agent experience can be abstracted into reusable procedural skills.}
During acquisition, each eligible attempt produces logged execution artifacts that are compacted into a structured trajectory summary and paired with verifier feedback.
A host-side Skill Author call, separated from the task-solving agent loop, uses this evidence and the current family skill state to write, refine, or skip a library update.
The resulting skill library is frozen before harder deployment tasks, so later success depends on whether prior experience has already been converted into reusable procedure rather than direct trace replay, continued adaptation, or test-time repair.
}
    \label{fig:teaser}
    \vspace{-10pt}
\end{figure*}
\section{Introduction}
\label{sec:intro}

Large language model (LLM) agents are increasingly being deployed as practical interfaces for real-world tasks. 
Unlike static question-answering systems, these agents interact with external environments over multi-step trajectories by reasoning, calling tools, inspecting files, executing code, and observing feedback~\cite{yao2023react,schick2023toolformer,zhou2024webarena,jimenez2024swebench}. 
As agents act over longer horizons, each task attempt leaves behind an episodic trajectory that records how the attempt unfolded. 
Prior work has shown that such experience can be stored and reused in later tasks~\cite{shinn2023reflexion,zhao2024expel,zheng2024synapse,wang2025agent}. 
Yet reusing an episode is not the same as extracting a procedure. 
A trajectory records what happened once and often mixes transferable decisions with incidental details, failed hypotheses, and mistakes. 
Future tasks rarely repeat the same episode exactly. 
They require a more explicit procedural form that states what to do again, when to do it, and what to check along the way. 
\textbf{\emph{Agent skills}}~\cite{agentskills_spec,anthropic2025agentskills} address this gap by turning reusable know-how into external artifacts that future agents can load, invoke when relevant, and follow across related tasks without replaying the original episode.

SkillsBench~\cite{li2026skillsbench} recently shows that curated skills improve agent performance across diverse domains, while self-generated skills provide little benefit on average. 
However, its self-generated setting is cold-start: agents write procedural guidance before attempting the task or observing verifier feedback.
This leaves open the central gap between \emph{skill use} and \emph{skill formation}. 
If curated skills show that procedural knowledge is useful, and experience-reuse methods show that trajectories contain task-solving evidence, can agents distill noisy one-off experience into compact skills that future agents can load, follow, and apply beyond the original episode, instead of merely replaying a trace?

To study this question, we introduce \textbf{\skillevolbench}, a diagnostic benchmark for the missing step between episodic experience and procedural reuse.
As shown in~\Cref{fig:teaser}, the benchmark turns each learning attempt into an abstraction step: an episodic trajectory and structured verifier feedback are passed to a host-side Skill Author call, which decides whether to write a new skill, revise an existing one, or leave the library unchanged. 
The resulting library is then frozen before harder related tasks are evaluated, so success depends on whether noisy one-off experience has already been encoded as reusable procedure. 
\skillevolbench~spans real-world work from engineering workflows to information work and workplace operations.
It contains six environments, each with five task families, where each family shares a latent procedural pattern across related problems.
Within each family, three learning tasks move from a canonical episode to targeted variants that expose the limits of a naive procedure, and three frozen evaluation tasks test transfer under context shift, adversarial shortcuts, and multi-skill composition.
Performance therefore reflects whether the agent has extracted what should generalize from noisy episodes before harder related tasks are seen.

We evaluate this question in both \textsc{Self-Generated} and \textsc{Curated-Start} settings.
The \textsc{Self-Generated} setting tests whether agents can induce skills from their own learning episodes, while the \textsc{Curated-Start} setting tests whether experience can improve human-written procedural priors.
We compare both against \textsc{No-Skill} and \textsc{Raw-Trajectory} controls, and replay the original learning tasks with the final frozen library to separate local recovery from deployment transfer.

This design makes three aspects of skill evolution measurable.
First, \skillevolbench~evaluates skill formation rather than only skill use: each family requires agents to convert verifier-grounded episodes into a persistent procedural artifact before harder related tasks are seen.
Second, its role-conditioned task arcs separate acquisition, replay, transfer, context shift, adversarial robustness, and composition, exposing failure modes that a single success rate would hide.
Third, its controls distinguish procedural abstraction from base agent capability, curated prior knowledge, and direct reuse of raw episodic traces.

Across ten model configurations and three agent harnesses, we find that current agents exhibit local procedural adaptation but not reliable reusable skill formation.
Skill-based agents can improve acquisition or replay, yet these gains do not consistently transfer to frozen deployment tasks.
\textsc{Raw-Trajectory} controls reveal a lossy abstraction bottleneck: agents often use episodic traces more effectively than the distilled skills derived from them.
Additional diagnostics show that the bottleneck is not simply capacity or cost: larger resource libraries and more frequent authoring can help in isolated cases, but they also introduce episode-specific drift, procedural clutter, and model-dependent failures.
Together, these results position \skillevolbench~as a diagnostic testbed for studying when experience becomes a reusable skill, when it remains an episode-specific patch, and when skill abstraction loses information needed for future tasks.
\section{Related Work}
\label{sec:related-work}

\textbf{\noindent{From static tasks to realistic agent work.}}
Agent benchmarks have increasingly moved from static tasks toward interactive settings that resemble real-world work~\cite{trivedi2024appworld,merrill2026terminal}. 
Mind2Web, MindWeb, and WebArena evaluate multi-step web navigation~\cite{deng2023mind2web,gou2026mindweb,zhou2024webarena}; SWE-bench grounds software-engineering evaluation in real GitHub issues~\cite{jimenez2024swebench}; and OSWorld, $\tau$-bench, and TheAgentCompany extend evaluation to computer use, user interaction, tool policies, and workplace workflows~\cite{xie2024osworld,yao2025taubench,xu2026theagentcompany}. 
These benchmarks make agent evaluation more realistic, but they mainly measure whether an agent can complete a task rather than whether its experience becomes a reusable procedure for later related tasks.

\textbf{\noindent{Reusing agent experience.}}
A growing line of work studies how agents improve by reusing prior experience without updating model parameters. 
Reflexion stores verbal feedback in episodic memory~\cite{shinn2023reflexion}, ExpeL extracts lessons from accumulated experiences~\cite{zhao2024expel}, Synapse retrieves complete past trajectories as exemplars~\cite{zheng2024synapse}, and Agent Workflow Memory induces reusable workflows from web-agent executions~\cite{wang2025agent}. 
These methods show that trajectories and reflections contain useful task-solving evidence, but they primarily reuse episodic traces or derived lessons rather than evaluating whether such evidence becomes durable procedural artifacts.

\textbf{\noindent{Agent Skills and skill evolution.}}
Agent Skills make procedural knowledge explicit by packaging task guidance, scripts, references, and resources into loadable artifacts~\cite{anthropic2025agentskills,agentskills_spec}. 
SkillsBench shows that curated skills can improve performance, while cold-start self-generated skills provide limited average gains~\cite{li2026skillsbench}. 
Related systems study LLM-generated tools~\cite{qian2023creator,cai2024large}, executable code-skill libraries~\cite{wang2024voyager}, and skill discovery, memory skills, self-evolution, or trajectory-derived skill libraries~\cite{xia2026skillrl,yang2026autoskill,zhang2026memskill,zhou2026memento,zhang2026evoskills,ma2026skillclaw,chen2026skillcraft,alzubi2026evoskill,wang2026skillx}. 
SkillEvolBench complements this work by testing whether verifier-grounded task episodes can yield external skill artifacts that persist under frozen deployment, context shift, adversarial shortcuts, and multi-skill composition.
\section{\skillevolbench}
\label{sec:skill-evol-bench}

\subsection{Overview}
\label{sec:overview}

SkillEvolBench evaluates whether agents can transform repeated task experience into reusable procedural skills. 
It contains 180 tasks across six real-world agent environments, with five task families per environment and six role-conditioned tasks per family. 
Each family defines a skill-evolution arc: tasks share an underlying procedure but vary failure modes, surface forms, and deployment conditions. This design distinguishes task-specific fixes from skills that can be revised, invoked, and composed.


\subsection{Environments and Task Families}
\label{sec:task_families}

\begin{figure*}[t]
    \centering
    \vspace{-20pt}
    \includegraphics[width=\linewidth]{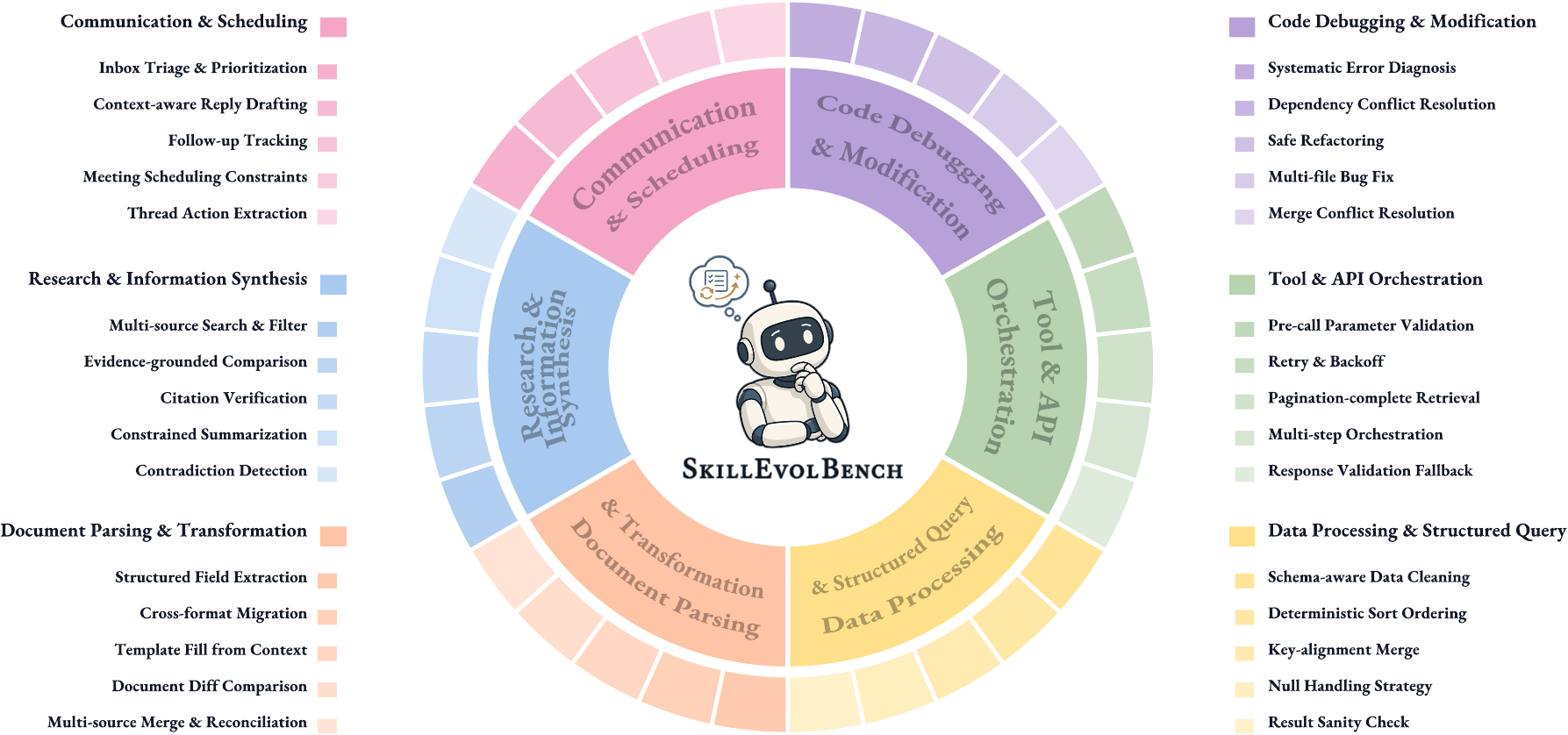}
    \caption{
    \textbf{\textsc{SkillEvolBench} organizes agent work into controlled skill-evolution arcs.}
    The benchmark spans 180 tasks across six real-world environments, with five recurring procedural families in each environment.
    Each family follows the same six-role progression: canonical, enriched, and variant tasks expose and stress-test the target procedure, while context-shift, adversarial, and composition tasks evaluate whether the evolved skill transfers, resists shortcuts, and combines with other skills in realistic workflows.
    }
    \label{fig:taxonomy}
    \vspace{-10pt}
\end{figure*}

\Cref{fig:taxonomy} summarizes the taxonomy. 
The six environments cover common forms of agent work: code modification, API orchestration, data processing, document transformation, research synthesis, and communication operations. 
A task family denotes a recurring procedural capability rather than a topic label, so families are related enough for experience to matter but varied enough to separate procedural learning from memorized topics.

\subsection{Task Construction}
\label{sec:task_construction}

\textbf{\noindent{Task selection.}}
We construct SkillEvolBench through a source-driven and human-curated process. 
We do not reuse existing benchmark instances. 
Instead, we use open-source agent skill collections, skill-oriented benchmarks, and practitioner-facing examples as evidence for task topics and workflow motifs~\citep{li2026skillsbench,han2026swe,kiloai2026pinchbench,anthropic2025agentskills,anthropic2025skillsrepo}. 
These sources guide the design space but do not define the tasks directly. 
We cluster observed workflows by artifact type, required tools, interaction pattern, and solution procedure, then retain families that satisfy three desiderata: real-world relevance, procedural skill fit, and verifiable evolvability. 
In particular, each family must describe a reusable procedure that is specific enough to be written as a skill, general enough to transfer beyond one fixture, and evaluable through deterministic outcome checks and process-level evidence.

\textbf{\noindent{Role-conditioned progression.}}
For each family, we instantiate six roles. 
The first three support skill acquisition: the \emph{canonical} task presents the base procedure, the \emph{enriched} task exposes a missing sub-capability, and the \emph{variant} task changes the surface form while preserving the same procedure. 
The last three evaluate deployment: the \emph{context-shift} task embeds the skill need in a broader request, the \emph{adversarial} task introduces shortcut solutions that can pass shallow checks, and the \emph{composition} task requires the target skill to interact with other skills. 
This progression tests family-level transfer, implicit invocation, shortcut resistance, and composition.

\textbf{\noindent{Gap-exposed curated skills.}}
For each family, we provide a \emph{gap-exposed curated skill}. 
It is neither an oracle solution nor copied from any task instance. 
We first define the family-level procedure and the gaps that should remain exposed. 
A \texttt{skill-creator} drafts an initial skill from this specification~\citep{anthropic2026skillcreator}, and we manually refine the draft to control its granularity. 
The resulting skill should support the canonical task but leaves enriched, variant, adversarial, and compositional cases unresolved; curated-start
agents therefore receive a useful but bounded initialization that still requires experience-driven refinement.

\textbf{\noindent{Specification and review.}}
Each task contains an instruction and fixture, a verification suite, and a scoring rubric. 
The verification suite includes public tests for the basic contract, hidden tests for edge cases and distribution shifts, and process verifiers that inspect traces and artifacts for brittle strategies such as hard-coded constants, swallowed exceptions, skipped validation, or incomplete repairs. 
Before inclusion, each family and curated skill is manually reviewed for realism, role alignment, verifier coverage, and whether the curated skill is useful but incomplete. 
The complete specifications are provided in Appendix~\Cref{app:family_catalog,app:task_design_catalog}.

\section{Skill Evolution Protocol}
\label{sec:skill_evolution_protocol}

\subsection{Protocol Overview}
\label{sec:protocol_overview}

\begin{figure*}[t]
    \centering
    \includegraphics[width=\textwidth]{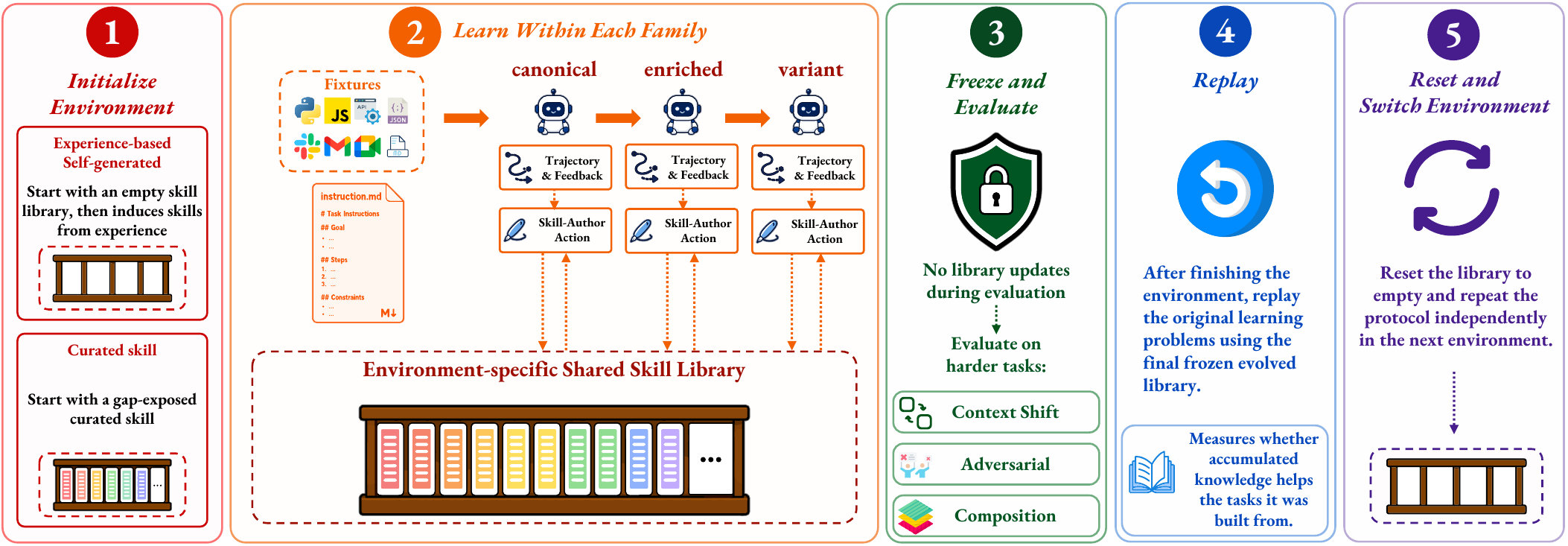}
    \vspace{-10pt}
    \caption{
    \textbf{\textsc{SkillEvolBench} evaluation protocol.}
    Each environment forms a self-contained lifelong episode.
    Agents learn from canonical, enriched, and variant tasks, using trajectories and verifier feedback to update an environment-specific skill library.
    The library is then frozen before context-shift, adversarial, and composition evaluation, so deployment success depends on prior skill formation rather than test-time repair.
    Replay measures whether the final frozen library also improves the original learning tasks, separating local recovery from transfer to harder roles.
    After replay, the library is reset before the next environment to prevent cross-environment leakage.
    }
    \label{fig:protocol}
    \vspace{-10pt}
\end{figure*}

Each environment in SkillEvolBench is evaluated as an independent lifelong episode. As shown in \Cref{fig:protocol}, an episode activates a fresh environment-scoped skill library. The agent first completes acquisition tasks, where logged execution artifacts are compacted and paired with verifier feedback as evidence for possible skill updates; implementation details of trajectory compaction are provided in Appendix~\Cref{app:trajectory_compactor}. The resulting library is then frozen for deployment. When enabled, replay reruns the original acquisition tasks with the final frozen library. Moving to a new environment activates a fresh library; skills from previous environments are retained only for logging and audit and are not mounted for the next episode.

\subsection{Initialization and Skill Conditions}
\label{sec:skill_conditions}

We compare three ways of initializing family-level procedural knowledge. In the \emph{experience-based self-generated} condition, a family starts with no skill; the \emph{canonical} task is attempted without a family skill, and induction may occur only after execution evidence and verifier feedback are available. In the \emph{zero-shot generated} condition, a metadata-only skill is generated before execution and remains fixed. In the \emph{curated} condition, a family starts from a \emph{gap-exposed curated skill}, which covers the base procedure but leaves room for acquisition tasks to expose missing sub-capabilities. The curated seed is fixed in static variants and may be refined only when revision is enabled.

For environment \(e\) and family \(f\), the starting skill set under condition \(c\) is
\begin{equation}
S_c(e,f)=
\begin{cases}
\emptyset, & c=\mathrm{exp\mbox{-}self},\\
\{\hat{s}^{0}_{e,f}\}, & c=\mathrm{zero\mbox{-}shot},\\
\{s^{\mathrm{gap}}_{e,f}\}, & c=\mathrm{curated}.
\end{cases}
\label{eq:starting_skill}
\end{equation}
Here, \(\hat{s}^{0}_{e,f}\) denotes the zero-shot skill, and
\(s^{\mathrm{gap}}_{e,f}\) denotes the gap-exposed curated seed defined
in \Cref{sec:task_construction}; \Cref{app:skill_prompts} gives the
authoring prompts for zero-shot generation, experience-based induction,
and revision.

\subsection{Acquisition: From Episodic Evidence to Skill Updates}
\label{sec:acquisition}

Let \(x_{e,f}^{r}\) denote the task in environment \(e\), family \(f\),
and role \(r\). We write the three acquisition roles as
\begin{equation}
\mathcal{R}_{\mathrm{acq}}=\{\mathrm{can},\mathrm{enr},\mathrm{var}\},
\label{eq:acq_roles}
\end{equation}
where \(\mathrm{can}\), \(\mathrm{enr}\), and \(\mathrm{var}\) denote the
canonical, enriched, and variant learning tasks.

All acquisition tasks in an environment are completed before deployment begins. The active library is scoped to the environment, so skills learned from earlier families may be visible to later families in the same environment, but never transfer across environments. A family-level starting skill \(S_c(e,f)\), when present, is introduced when that family's \emph{canonical} task is first reached.

Each acquisition attempt yields a compacted trajectory summary \(\tilde{\tau}_{e,f}^{r}\) from harness-recorded artifacts such as instructions, file accesses, tool calls, commands, edits, generated outputs, tests, and final responses. The verifier returns feedback \(v_{e,f}^{r}\), including outcome results, process checks, rewards, and diagnostics. We do not access hidden model state or private chain-of-thought.

Skill authoring is family-local. Although the task-solving agent may read the environment-level library, the Skill Author receives only same-family skills and same-family acquisition history. With \(\mathrm{can}\prec\mathrm{enr}\prec\mathrm{var}\), the available evidence after role \(r\) is
\begin{equation}
H_{e,f}^{\preceq r}
=
\bigl((\tilde{\tau}_{e,f}^{r'},v_{e,f}^{r'})\bigr)_
{r'\in\mathcal{R}_{\mathrm{acq}},\,r'\preceq r}.
\label{eq:family_history}
\end{equation}
The \texttt{Skill Author} is invoked only after eligible acquisition attempts and emits a structured library edit:
\begin{equation}
L_{e,k+1}=U_c\!\left(L_{e,k}, H_{e,f_k}^{\preceq r_k}\right).
\label{eq:skill_update}
\end{equation}
\begin{table*}[t]
\centering
\caption{
\textbf{\textsc{No-Skill} vs. \textsc{Curated-Start} skills.}
Values are success rates (\%). 
Deltas report percentage-point differences from the corresponding \textsc{No-Skill} result.
For RSR, the reference is \textsc{No-Skill} LSR because \textsc{No-Skill} has no replay phase.
\textcolor{gainred}{Red}/\textcolor{lossblue}{blue}/\textcolor{neutralgray}{gray} denote positive/negative/zero deltas.
}
\label{tab:noskill_vs_curated_compact}
\vspace{0.35em}
\scriptsize
\setlength{\tabcolsep}{1.6pt}
\renewcommand{\arraystretch}{1.10}
\begin{adjustbox}{max width=\linewidth}
\begin{tabular}{@{}cclrrrrrrrrrrrr@{}}
\toprule
\textbf{Agent Harness}
& \textbf{Model}
& \textbf{Condition}
& \multicolumn{2}{c}{\textbf{LSR}}
& \multicolumn{2}{c}{\textbf{RSR}}
& \multicolumn{2}{c}{\textbf{ESR}}
& \multicolumn{2}{c}{\textbf{CSSR}}
& \multicolumn{2}{c}{\textbf{ARSR}}
& \multicolumn{2}{c}{\textbf{CompSR}} \\
\cmidrule(lr){4-5}
\cmidrule(lr){6-7}
\cmidrule(lr){8-9}
\cmidrule(lr){10-11}
\cmidrule(lr){12-13}
\cmidrule(lr){14-15}
& &
& \textbf{\%} & \(\Delta\)
& \textbf{\%} & \(\Delta\)
& \textbf{\%} & \(\Delta\)
& \textbf{\%} & \(\Delta\)
& \textbf{\%} & \(\Delta\)
& \textbf{\%} & \(\Delta\) \\
\midrule
\harnesscell{16}{\textsc{Claude Code}} & \multirow{4}{*}{Claude Opus 4.6} & \cond{No-Skill} & 38.9 & \NA & -- & \NA & 37.8 & \NA & 46.7 & \NA & 36.7 & \NA & 30.0 & \NA \\
 &  & \cond{Curated-Static} & 40.0 & \pos{1.1} & -- & \NA & 34.4 & \down{-3.4} & 46.7 & \zero{0.0} & 36.7 & \zero{0.0} & 20.0 & \down{-10.0} \\
 &  & \cond{~~~~+ Revision} & 44.4 & \pos{5.5} & 56.7 & \pos{17.8} & 38.9 & \pos{1.1} & 40.0 & \down{-6.7} & 43.3 & \pos{6.6} & 33.3 & \pos{3.3} \\
 &  & \cond{~~~~+ Always} & 42.2 & \pos{3.3} & 46.7 & \pos{7.8} & 37.8 & \zero{0.0} & 40.0 & \down{-6.7} & 46.7 & \pos{10.0} & 26.7 & \down{-3.3} \\
\addlinespace[0.10em]
\cdashline{2-15}
\addlinespace[0.10em]
 & \multirow{4}{*}{Claude Opus 4.5} & \cond{No-Skill} & 42.2 & \NA & -- & \NA & 32.2 & \NA & 40.0 & \NA & 40.0 & \NA & 16.7 & \NA \\
 &  & \cond{Curated-Static} & 45.6 & \pos{3.4} & -- & \NA & 34.4 & \pos{2.2} & 43.3 & \pos{3.3} & 40.0 & \zero{0.0} & 20.0 & \pos{3.3} \\
 &  & \cond{~~~~+ Revision} & 38.9 & \down{-3.3} & 42.2 & \zero{0.0} & 34.4 & \pos{2.2} & 43.3 & \pos{3.3} & 36.7 & \down{-3.3} & 23.3 & \pos{6.6} \\
 &  & \cond{~~~~+ Always} & 42.2 & \zero{0.0} & 44.4 & \pos{2.2} & 36.7 & \pos{4.5} & 43.3 & \pos{3.3} & 43.3 & \pos{3.3} & 23.3 & \pos{6.6} \\
\addlinespace[0.10em]
\cdashline{2-15}
\addlinespace[0.10em]
 & \multirow{4}{*}{Claude Sonnet 4.6} & \cond{No-Skill} & 37.8 & \NA & -- & \NA & 38.9 & \NA & 40.0 & \NA & 50.0 & \NA & 26.7 & \NA \\
 &  & \cond{Curated-Static} & 41.1 & \pos{3.3} & -- & \NA & 35.6 & \down{-3.3} & 36.7 & \down{-3.3} & 46.7 & \down{-3.3} & 23.3 & \down{-3.4} \\
 &  & \cond{~~~~+ Revision} & 41.1 & \pos{3.3} & 51.1 & \pos{13.3} & 38.9 & \zero{0.0} & 46.7 & \pos{6.7} & 43.3 & \down{-6.7} & 26.7 & \zero{0.0} \\
 &  & \cond{~~~~+ Always} & 40.0 & \pos{2.2} & 44.4 & \pos{6.6} & 38.9 & \zero{0.0} & 46.7 & \pos{6.7} & 43.3 & \down{-6.7} & 26.7 & \zero{0.0} \\
\addlinespace[0.10em]
\cdashline{2-15}
\addlinespace[0.10em]
 & \multirow{4}{*}{Claude Sonnet 4.5} & \cond{No-Skill} & 41.1 & \NA & -- & \NA & 35.6 & \NA & 40.0 & \NA & 46.7 & \NA & 20.0 & \NA \\
 &  & \cond{Curated-Static} & 35.6 & \down{-5.5} & -- & \NA & 36.7 & \pos{1.1} & 43.3 & \pos{3.3} & 43.3 & \down{-3.4} & 23.3 & \pos{3.3} \\
 &  & \cond{~~~~+ Revision} & 40.0 & \down{-1.1} & 42.2 & \pos{1.1} & 28.9 & \down{-6.7} & 36.7 & \down{-3.3} & 33.3 & \down{-13.4} & 16.7 & \down{-3.3} \\
 &  & \cond{~~~~+ Always} & 37.8 & \down{-3.3} & 38.9 & \down{-2.2} & 33.3 & \down{-2.3} & 40.0 & \zero{0.0} & 40.0 & \down{-6.7} & 20.0 & \zero{0.0} \\
 
\midrule

\harnesscell{12}{\textsc{Codex CLI}} & \multirow{4}{*}{GPT-5.4} & \cond{No-Skill} & 43.3 & \NA & -- & \NA & 33.3 & \NA & 43.3 & \NA & 33.3 & \NA & 23.3 & \NA \\
 &  & \cond{Curated-Static} & 43.3 & \zero{0.0} & -- & \NA & 32.2 & \down{-1.1} & 40.0 & \down{-3.3} & 40.0 & \pos{6.7} & 16.7 & \down{-6.6} \\
 &  & \cond{~~~~+ Revision} & 45.6 & \pos{2.3} & 45.6 & \pos{2.3} & 38.9 & \pos{5.6} & 50.0 & \pos{6.7} & 43.3 & \pos{10.0} & 23.3 & \zero{0.0} \\
 &  & \cond{~~~~+ Always} & 43.3 & \zero{0.0} & 42.2 & \down{-1.1} & 40.0 & \pos{6.7} & 40.0 & \down{-3.3} & 50.0 & \pos{16.7} & 30.0 & \pos{6.7} \\
\addlinespace[0.10em]
\cdashline{2-15}
\addlinespace[0.10em]
 & \multirow{4}{*}{GPT-5.3-Codex} & \cond{No-Skill} & 44.4 & \NA & -- & \NA & 34.4 & \NA & 36.7 & \NA & 46.7 & \NA & 20.0 & \NA \\
 &  & \cond{Curated-Static} & 45.6 & \pos{1.2} & -- & \NA & 32.2 & \down{-2.2} & 36.7 & \zero{0.0} & 43.3 & \down{-3.4} & 16.7 & \down{-3.3} \\
 &  & \cond{~~~~+ Revision} & 42.2 & \down{-2.2} & 48.9 & \pos{4.5} & 34.4 & \zero{0.0} & 36.7 & \zero{0.0} & 46.7 & \zero{0.0} & 20.0 & \zero{0.0} \\
 &  & \cond{~~~~+ Always} & 42.2 & \down{-2.2} & 45.6 & \pos{1.2} & 32.2 & \down{-2.2} & 40.0 & \pos{3.3} & 40.0 & \down{-6.7} & 16.7 & \down{-3.3} \\
\addlinespace[0.10em]
\cdashline{2-15}
\addlinespace[0.10em]
 & \multirow{4}{*}{GPT-5.2-Codex} & \cond{No-Skill} & 43.3 & \NA & -- & \NA & 36.7 & \NA & 40.0 & \NA & 53.3 & \NA & 16.7 & \NA \\
 &  & \cond{Curated-Static} & 51.1 & \pos{7.8} & -- & \NA & 30.0 & \down{-6.7} & 40.0 & \zero{0.0} & 36.7 & \down{-16.6} & 13.3 & \down{-3.4} \\
 &  & \cond{~~~~+ Revision} & 45.6 & \pos{2.3} & 43.3 & \zero{0.0} & 36.7 & \zero{0.0} & 43.3 & \pos{3.3} & 46.7 & \down{-6.6} & 20.0 & \pos{3.3} \\
 &  & \cond{~~~~+ Always} & 44.4 & \pos{1.1} & 45.6 & \pos{2.3} & 37.8 & \pos{1.1} & 46.7 & \pos{6.7} & 43.3 & \down{-10.0} & 23.3 & \pos{6.6} \\
\addlinespace[0.18em]
\midrule
\harnesscell{12}{\textsc{Gemini CLI}} & \multirow{4}{*}{Gemini 3.1 Pro} & \cond{No-Skill} & 40.0 & \NA & -- & \NA & 35.6 & \NA & 40.0 & \NA & 46.7 & \NA & 20.0 & \NA \\
 &  & \cond{Curated-Static} & 44.4 & \pos{4.4} & -- & \NA & 35.6 & \zero{0.0} & 36.7 & \down{-3.3} & 43.3 & \down{-3.4} & 26.7 & \pos{6.7} \\
 &  & \cond{~~~~+ Revision} & 43.3 & \pos{3.3} & 55.6 & \pos{15.6} & 35.6 & \zero{0.0} & 36.7 & \down{-3.3} & 43.3 & \down{-3.4} & 26.7 & \pos{6.7} \\
 &  & \cond{~~~~+ Always} & 40.0 & \zero{0.0} & 45.6 & \pos{5.6} & 35.6 & \zero{0.0} & 40.0 & \zero{0.0} & 40.0 & \down{-6.7} & 26.7 & \pos{6.7} \\
\addlinespace[0.10em]
\cdashline{2-15}
\addlinespace[0.10em]
 & \multirow{4}{*}{Gemini 3 Flash} & \cond{No-Skill} & 40.0 & \NA & -- & \NA & 35.6 & \NA & 30.0 & \NA & 53.3 & \NA & 23.3 & \NA \\
 &  & \cond{Curated-Static} & 41.1 & \pos{1.1} & -- & \NA & 32.2 & \down{-3.4} & 36.7 & \pos{6.7} & 43.3 & \down{-10.0} & 16.7 & \down{-6.6} \\
 &  & \cond{~~~~+ Revision} & 42.2 & \pos{2.2} & 42.2 & \pos{2.2} & 32.2 & \down{-3.4} & 30.0 & \zero{0.0} & 46.7 & \down{-6.6} & 20.0 & \down{-3.3} \\
 &  & \cond{~~~~+ Always} & 43.3 & \pos{3.3} & 37.8 & \down{-2.2} & 37.8 & \pos{2.2} & 46.7 & \pos{16.7} & 36.7 & \down{-16.6} & 30.0 & \pos{6.7} \\
\addlinespace[0.10em]
\cdashline{2-15}
\addlinespace[0.10em]
 & \multirow{4}{*}{Gemini 2.5 Pro} & \cond{No-Skill} & 30.0 & \NA & -- & \NA & 26.7 & \NA & 26.7 & \NA & 30.0 & \NA & 23.3 & \NA \\
 &  & \cond{Curated-Static} & 30.0 & \zero{0.0} & -- & \NA & 18.9 & \down{-7.8} & 20.0 & \down{-6.7} & 23.3 & \down{-6.7} & 13.3 & \down{-10.0} \\
 &  & \cond{~~~~+ Revision} & 28.9 & \down{-1.1} & 32.2 & \pos{2.2} & 21.1 & \down{-5.6} & 16.7 & \down{-10.0} & 30.0 & \zero{0.0} & 16.7 & \down{-6.6} \\
 &  & \cond{~~~~+ Always} & 31.1 & \pos{1.1} & 26.7 & \down{-3.3} & 24.4 & \down{-2.3} & 26.7 & \zero{0.0} & 23.3 & \down{-6.7} & 23.3 & \zero{0.0} \\
\bottomrule
\end{tabular}
\end{adjustbox}
\vspace{-10pt}
\end{table*}

The update rule \(U_c\) depends on the condition. Experience-based self-generation may induce a new skill after the \emph{canonical} attempt and revise it on later failed acquisition attempts. Always-update variants invoke authoring after every eligible acquisition attempt. Curated revision variants refine the curated seed under the same trigger policy, while curated static keeps it fixed. Zero-shot skills are never revised:
\begin{equation}
U_{\mathrm{zero}}\!\left(L,H\right)=L.
\label{eq:zero_no_authoring}
\end{equation}
\subsection{Frozen Deployment and Replay}
\label{sec:frozen_replay}

After acquisition, the environment-specific library is frozen. Deployment uses the \emph{context-shift}, \emph{adversarial}, and \emph{composition} roles. During deployment, the agent may read and apply accumulated skills, but may not create, revise, retire, or otherwise modify the library. This phase measures whether prior skill evolution transfers to harder tasks without allowing adaptation on the evaluation instance itself. When replay is enabled, we rerun the original acquisition tasks using the final frozen library. Replay does not update the library. It provides a within-environment counterfactual: the same learning tasks are solved once before the relevant skills have matured and once after the library has evolved.
\begin{table*}[t]
\centering
\caption{
\textbf{\textsc{No-Skill} vs. \textsc{Self-Generated} skills.}
Values are success rates (\%) with percentage-point deltas relative to \textsc{No-Skill}.
RSR deltas use \textsc{No-Skill} LSR as the baseline.
Zero-shot skills are metadata-induced and unrevised.
}
\label{tab:noskill_vs_selfgen_compact}
\vspace{0.35em}
\scriptsize
\setlength{\tabcolsep}{1.6pt}
\renewcommand{\arraystretch}{1.10}
\begin{adjustbox}{max width=\linewidth}
\begin{tabular}{@{}cclrrrrrrrrrrrr@{}}
\toprule
\textbf{Agent Harness}
& \textbf{Model}
& \textbf{Condition}
& \multicolumn{2}{c}{\textbf{LSR}}
& \multicolumn{2}{c}{\textbf{RSR}}
& \multicolumn{2}{c}{\textbf{ESR}}
& \multicolumn{2}{c}{\textbf{CSSR}}
& \multicolumn{2}{c}{\textbf{ARSR}}
& \multicolumn{2}{c}{\textbf{CompSR}} \\
\cmidrule(lr){4-5}
\cmidrule(lr){6-7}
\cmidrule(lr){8-9}
\cmidrule(lr){10-11}
\cmidrule(lr){12-13}
\cmidrule(lr){14-15}
& &
& \textbf{\%} & \(\Delta\)
& \textbf{\%} & \(\Delta\)
& \textbf{\%} & \(\Delta\)
& \textbf{\%} & \(\Delta\)
& \textbf{\%} & \(\Delta\)
& \textbf{\%} & \(\Delta\) \\
\midrule
\harnesscell{16}{\textsc{Claude Code}} & \multirow{4}{*}{Claude Opus 4.6} & \cond{No-Skill} & 38.9 & \NA & -- & \NA & 37.8 & \NA & 46.7 & \NA & 36.7 & \NA & 30.0 & \NA \\
 &  & \cond{Zero-Shot} & 41.1 & \pos{2.2} & -- & \NA & 32.2 & \down{-5.6} & 40.0 & \down{-6.7} & 33.3 & \down{-3.4} & 23.3 & \down{-6.7} \\
 &  & \cond{Experience} & 44.4 & \pos{5.5} & 48.9 & \pos{10.0} & 32.2 & \down{-5.6} & 40.0 & \down{-6.7} & 36.7 & \zero{0.0} & 20.0 & \down{-10.0} \\
 &  & \cond{~~~~+ Always} & 42.2 & \pos{3.3} & 45.6 & \pos{6.7} & 37.8 & \zero{0.0} & 43.3 & \down{-3.4} & 46.7 & \pos{10.0} & 23.3 & \down{-6.7} \\
\addlinespace[0.10em]
\cdashline{2-15}
\addlinespace[0.10em]
 & \multirow{4}{*}{Claude Opus 4.5} & \cond{No-Skill} & 42.2 & \NA & -- & \NA & 32.2 & \NA & 40.0 & \NA & 40.0 & \NA & 16.7 & \NA \\
 &  & \cond{Zero-Shot} & 41.1 & \down{-1.1} & -- & \NA & 34.4 & \pos{2.2} & 43.3 & \pos{3.3} & 40.0 & \zero{0.0} & 20.0 & \pos{3.3} \\
 &  & \cond{Experience} & 42.2 & \zero{0.0} & 40.0 & \down{-2.2} & 31.1 & \down{-1.1} & 33.3 & \down{-6.7} & 40.0 & \zero{0.0} & 20.0 & \pos{3.3} \\
 &  & \cond{~~~~+ Always} & 41.1 & \down{-1.1} & 40.0 & \down{-2.2} & 35.6 & \pos{3.4} & 40.0 & \zero{0.0} & 40.0 & \zero{0.0} & 26.7 & \pos{10.0} \\
\addlinespace[0.10em]
\cdashline{2-15}
\addlinespace[0.10em]
 & \multirow{4}{*}{Claude Sonnet 4.6} & \cond{No-Skill} & 37.8 & \NA & -- & \NA & 38.9 & \NA & 40.0 & \NA & 50.0 & \NA & 26.7 & \NA \\
 &  & \cond{Zero-Shot} & 37.8 & \zero{0.0} & -- & \NA & 36.7 & \down{-2.2} & 40.0 & \zero{0.0} & 43.3 & \down{-6.7} & 26.7 & \zero{0.0} \\
 &  & \cond{Experience} & 44.4 & \pos{6.6} & 46.7 & \pos{8.9} & 40.0 & \pos{1.1} & 46.7 & \pos{6.7} & 40.0 & \down{-10.0} & 33.3 & \pos{6.6} \\
 &  & \cond{~~~~+ Always} & 41.1 & \pos{3.3} & 46.7 & \pos{8.9} & 40.0 & \pos{1.1} & 40.0 & \zero{0.0} & 50.0 & \zero{0.0} & 30.0 & \pos{3.3} \\
\addlinespace[0.10em]
\cdashline{2-15}
\addlinespace[0.10em]
 & \multirow{4}{*}{Claude Sonnet 4.5} & \cond{No-Skill} & 41.1 & \NA & -- & \NA & 35.6 & \NA & 40.0 & \NA & 46.7 & \NA & 20.0 & \NA \\
 &  & \cond{Zero-Shot} & 41.1 & \zero{0.0} & -- & \NA & 27.8 & \down{-7.8} & 33.3 & \down{-6.7} & 30.0 & \down{-16.7} & 20.0 & \zero{0.0} \\
 &  & \cond{Experience} & 36.7 & \down{-4.4} & 41.1 & \zero{0.0} & 35.6 & \zero{0.0} & 40.0 & \zero{0.0} & 40.0 & \down{-6.7} & 26.7 & \pos{6.7} \\
 &  & \cond{~~~~+ Always} & 38.9 & \down{-2.2} & 44.4 & \pos{3.3} & 33.3 & \down{-2.3} & 40.0 & \zero{0.0} & 40.0 & \down{-6.7} & 20.0 & \zero{0.0} \\
\midrule
\harnesscell{12}{\textsc{Codex CLI}} & \multirow{4}{*}{GPT-5.4} & \cond{No-Skill} & 43.3 & \NA & -- & \NA & 33.3 & \NA & 43.3 & \NA & 33.3 & \NA & 23.3 & \NA \\
 &  & \cond{Zero-Shot} & 45.6 & \pos{2.3} & -- & \NA & 33.3 & \zero{0.0} & 43.3 & \zero{0.0} & 40.0 & \pos{6.7} & 16.7 & \down{-6.6} \\
 &  & \cond{Experience} & 45.6 & \pos{2.3} & 44.4 & \pos{1.1} & 31.1 & \down{-2.2} & 40.0 & \down{-3.3} & 36.7 & \pos{3.4} & 16.7 & \down{-6.6} \\
 &  & \cond{~~~~+ Always} & 44.4 & \pos{1.1} & 43.3 & \zero{0.0} & 35.6 & \pos{2.3} & 46.7 & \pos{3.4} & 40.0 & \pos{6.7} & 20.0 & \down{-3.3} \\
\addlinespace[0.10em]
\cdashline{2-15}
\addlinespace[0.10em]
 & \multirow{4}{*}{GPT-5.3-Codex} & \cond{No-Skill} & 44.4 & \NA & -- & \NA & 34.4 & \NA & 36.7 & \NA & 46.7 & \NA & 20.0 & \NA \\
 &  & \cond{Zero-Shot} & 46.7 & \pos{2.3} & -- & \NA & 34.4 & \zero{0.0} & 43.3 & \pos{6.6} & 40.0 & \down{-6.7} & 20.0 & \zero{0.0} \\
 &  & \cond{Experience} & 42.2 & \down{-2.2} & 48.9 & \pos{4.5} & 31.1 & \down{-3.3} & 40.0 & \pos{3.3} & 40.0 & \down{-6.7} & 13.3 & \down{-6.7} \\
 &  & \cond{~~~~+ Always} & 44.4 & \zero{0.0} & 48.9 & \pos{4.5} & 34.4 & \zero{0.0} & 36.7 & \zero{0.0} & 43.3 & \down{-3.4} & 23.3 & \pos{3.3} \\
\addlinespace[0.10em]
\cdashline{2-15}
\addlinespace[0.10em]
 & \multirow{4}{*}{GPT-5.2-Codex} & \cond{No-Skill} & 43.3 & \NA & -- & \NA & 36.7 & \NA & 40.0 & \NA & 53.3 & \NA & 16.7 & \NA \\
 &  & \cond{Zero-Shot} & 45.6 & \pos{2.3} & -- & \NA & 30.0 & \down{-6.7} & 33.3 & \down{-6.7} & 36.7 & \down{-16.6} & 20.0 & \pos{3.3} \\
 &  & \cond{Experience} & 45.6 & \pos{2.3} & 42.2 & \down{-1.1} & 34.4 & \down{-2.3} & 36.7 & \down{-3.3} & 43.3 & \down{-10.0} & 23.3 & \pos{6.6} \\
 &  & \cond{~~~~+ Always} & 46.7 & \pos{3.4} & 46.7 & \pos{3.4} & 38.9 & \pos{2.2} & 50.0 & \pos{10.0} & 43.3 & \down{-10.0} & 23.3 & \pos{6.6} \\
\midrule
\harnesscell{12}{\textsc{Gemini CLI}} & \multirow{4}{*}{Gemini 3.1 Pro} & \cond{No-Skill} & 40.0 & \NA & -- & \NA & 35.6 & \NA & 40.0 & \NA & 46.7 & \NA & 20.0 & \NA \\
 &  & \cond{Zero-Shot} & 43.3 & \pos{3.3} & -- & \NA & 36.7 & \pos{1.1} & 36.7 & \down{-3.3} & 46.7 & \zero{0.0} & 26.7 & \pos{6.7} \\
 &  & \cond{Experience} & 47.8 & \pos{7.8} & 53.3 & \pos{13.3} & 32.2 & \down{-3.4} & 36.7 & \down{-3.3} & 36.7 & \down{-10.0} & 23.3 & \pos{3.3} \\
 &  & \cond{~~~~+ Always} & 42.2 & \pos{2.2} & 41.1 & \pos{1.1} & 33.3 & \down{-2.3} & 33.3 & \down{-6.7} & 43.3 & \down{-3.4} & 23.3 & \pos{3.3} \\
\addlinespace[0.10em]
\cdashline{2-15}
\addlinespace[0.10em]
 & \multirow{4}{*}{Gemini 3 Flash} & \cond{No-Skill} & 40.0 & \NA & -- & \NA & 35.6 & \NA & 30.0 & \NA & 53.3 & \NA & 23.3 & \NA \\
 &  & \cond{Zero-Shot} & 38.9 & \down{-1.1} & -- & \NA & 40.0 & \pos{4.4} & 43.3 & \pos{13.3} & 53.3 & \zero{0.0} & 23.3 & \zero{0.0} \\
 &  & \cond{Experience} & 40.0 & \zero{0.0} & 43.3 & \pos{3.3} & 33.3 & \down{-2.3} & 36.7 & \pos{6.7} & 40.0 & \down{-13.3} & 23.3 & \zero{0.0} \\
 &  & \cond{~~~~+ Always} & 35.6 & \down{-4.4} & 40.0 & \zero{0.0} & 36.7 & \pos{1.1} & 36.7 & \pos{6.7} & 43.3 & \down{-10.0} & 30.0 & \pos{6.7} \\
\addlinespace[0.10em]
\cdashline{2-15}
\addlinespace[0.10em]
 & \multirow{4}{*}{Gemini 2.5 Pro} & \cond{No-Skill} & 30.0 & \NA & -- & \NA & 26.7 & \NA & 26.7 & \NA & 30.0 & \NA & 23.3 & \NA \\
 &  & \cond{Zero-Shot} & 28.9 & \down{-1.1} & -- & \NA & 15.6 & \down{-11.1} & 13.3 & \down{-13.4} & 13.3 & \down{-16.7} & 20.0 & \down{-3.3} \\
 &  & \cond{Experience} & 27.8 & \down{-2.2} & 32.2 & \pos{2.2} & 20.0 & \down{-6.7} & 20.0 & \down{-6.7} & 30.0 & \zero{0.0} & 10.0 & \down{-13.3} \\
 &  & \cond{~~~~+ Always} & 31.1 & \pos{1.1} & 34.4 & \pos{4.4} & 25.6 & \down{-1.1} & 40.0 & \pos{13.3} & 20.0 & \down{-10.0} & 16.7 & \down{-6.6} \\
\bottomrule
\end{tabular}
\end{adjustbox}
\end{table*}
\subsection{Scoring}
\label{sec:scoring}

For each task attempt \(a\), the verifier returns an outcome score \(O_a\), a process score \(P_a\), an overall score \(G_a\), and a binary success indicator \(B_a\). Outcome measures functional correctness through public and hidden tests, while process measures whether the agent followed the intended procedure. For any attempt set \(\mathcal{I}\), we report
\begin{equation}
\mathrm{SR}(\mathcal{I})=\frac{1}{|\mathcal{I}|}\sum_{a\in\mathcal{I}}B_a .
\label{eq:generic_sr}
\end{equation}
We instantiate \(\mathrm{SR}(\mathcal{I})\) on protocol-defined task subsets.
\(\mathrm{LSR}\) measures success on acquisition tasks, where the agent works
through the canonical, enriched, and variant roles while skill updates are still
allowed. \(\mathrm{RSR}\) measures replay success on the original acquisition
tasks after the environment library has been frozen, capturing local recovery
rather than transfer. \(\mathrm{ESR}\) measures frozen deployment success on
held-out context-shift, adversarial, and composition tasks, where the agent may
use but not update the final library. We decompose \(\mathrm{ESR}\) into
\(\mathrm{CSSR}\), \(\mathrm{ARSR}\), and \(\mathrm{CompSR}\), which measure
implicit skill invocation under context shift, robustness to shortcut solutions,
and multi-skill composition, respectively.
\section{Experimental Setup and Results}
\label{sec:experimental-setup-and-results}
\subsection{Agent Harnesses and Models}
\label{sec:agent_harnesses_models}
We evaluate SkillEvolBench with three agent harnesses: Claude Code~\cite{anthropic2026claudecode}, Codex CLI~\cite{openai2026codexcli}, and Gemini CLI~\cite{google2026geminicli}. We run all harnesses under the same benchmark protocol. We test ten model configurations across the three harnesses. Claude Code is evaluated with Opus 4.6, Opus 4.5, Sonnet 4.6, and Sonnet 4.5. Codex CLI is evaluated with GPT-5.4, GPT-5.3-Codex, and GPT-5.2-Codex. Gemini CLI is evaluated with Gemini 3.1 Pro, Gemini 3 Flash, and Gemini 2.5 Pro. 

\subsection{Experiment Variants}
\label{sec:experiment_variants}

We evaluate eight primary variants. \textsc{No-Skill} uses no persistent memory. \textsc{Raw-Trajectory} retrieves compacted same-family acquisition trajectories, without inducing procedural skills. \textsc{Curated-Static} provides fixed curated skills. \textsc{Curated-Revision} and \textsc{Curated-Revision-Always} start from curated skills and revise them after failed or all acquisition attempts, respectively. \textsc{SelfGen-Zero-Shot} generates fixed metadata-only skills before the \emph{canonical} task. \textsc{SelfGen-Revision} induces skills from \emph{canonical} trajectories and revises after failed later acquisition attempts. \textsc{SelfGen-Always} updates after every acquisition attempt. 

\begin{figure*}[t]
\centering
\includegraphics[width=\textwidth]{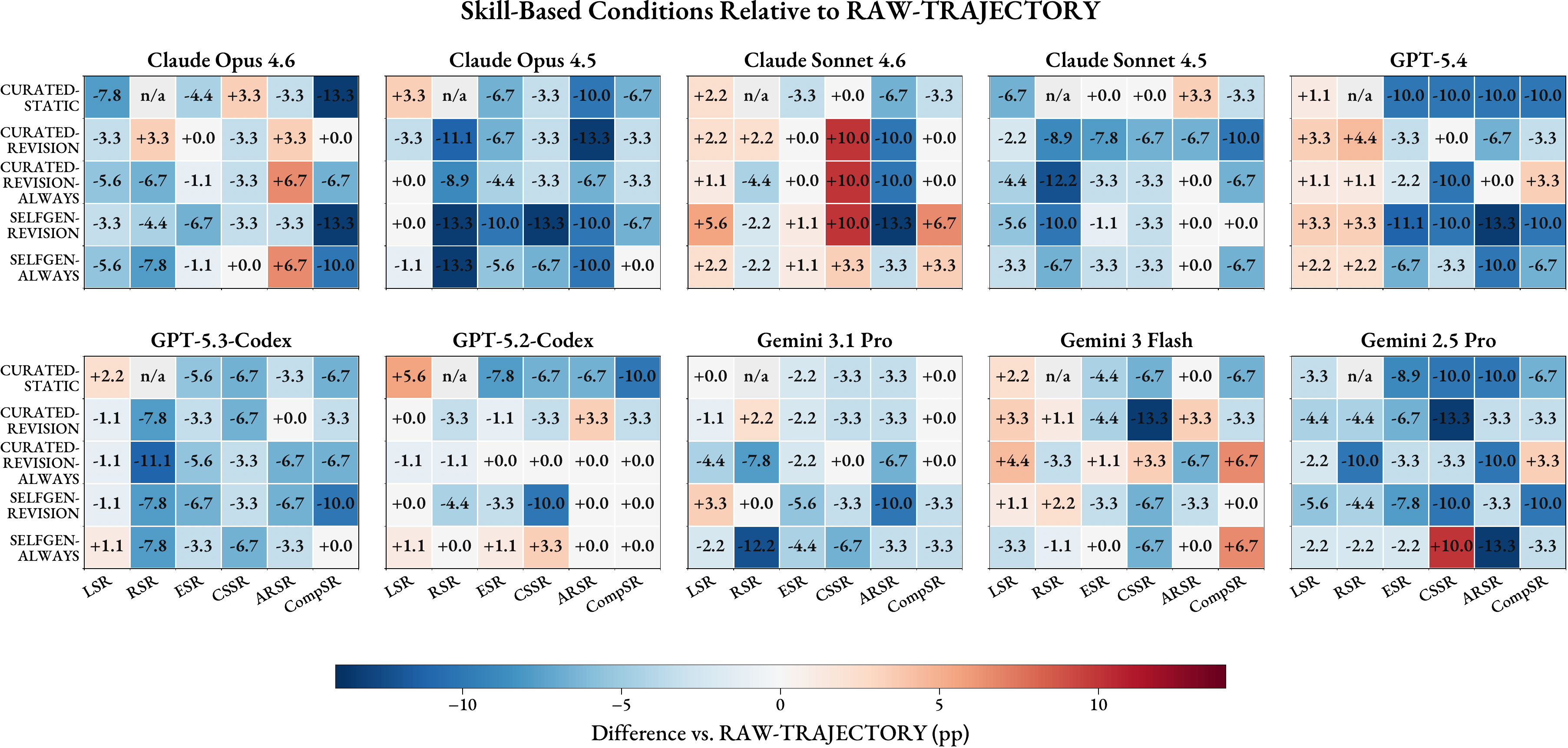}
\caption{
\textbf{Skill-based conditions relative to \textsc{Raw-Trajectory}.}
Each panel shows one model, with rows denoting skill variants and columns denoting success metrics.
Cells report percentage-point differences from the corresponding \textsc{Raw-Trajectory} baseline.
Positive values indicate that distilled skills outperform direct episodic reuse, while negative values indicate that raw trajectories preserve useful task evidence lost during skill abstraction.
Red indicates improvement, blue indicates degradation, and gray indicates unavailable comparisons.
}
\vspace{-10pt}
\label{fig:rawtraj_vs_skills_delta}
\end{figure*}

\subsection{Main Comparison: Does Episodic Experience Become Reusable Skills?}
\label{sec:main_results}

\textbf{\noindent{Overall observation.}}
\Cref{tab:noskill_vs_curated_compact,tab:noskill_vs_selfgen_compact} compare skill-based conditions against \textsc{No-Skill}, while \Cref{fig:rawtraj_vs_skills_delta} compares the same skill-based conditions against \textsc{Raw-Trajectory}. 
We interpret episodic experience as having become a reusable skill only when the resulting library improves not just the original acquisition or replay tasks, but also frozen deployment tasks that require invocation, robustness, and composition. 
Under this criterion, current agents exhibit local procedural adaptation but not reliable reusable skill formation. 
Skill-based conditions can improve LSR or RSR, and some model-condition pairs achieve strong gains on specific deployment axes. 
However, these gains do not consistently transfer across ESR, CSSR, ARSR, and CompSR. 
The \textsc{Raw-Trajectory} comparison further suggests a lossy abstraction bottleneck: agents often use episodic traces more effectively than the distilled skills derived from them.

\textbf{\noindent{Local gains do not imply reusable skill formation.}}
Both \textsc{Curated-Start} and \textsc{Self-Generated} settings show cases where skills improve LSR or RSR but fail on deployment metrics. 
For example, under \textsc{Self-Generated} \cond{Experience}, Claude Opus 4.6 improves LSR by 5.5 pp and RSR by 10.0 pp relative to \textsc{No-Skill}, but decreases ESR, CSSR, and CompSR. 
Claude Sonnet 4.6 under the same condition improves LSR, RSR, CSSR, and CompSR, yet drops by 10.0 pp on ARSR. 
Similarly, GPT-5.2-Codex with \textsc{Self-Generated} \cond{+ Always} improves most metrics but remains 10.0 pp below \textsc{No-Skill} on ARSR. 
These examples suggest that replay recovery and local improvement can reflect useful patches without guaranteeing robust procedural reuse.

\textbf{\noindent{Deployment metrics expose distinct failure modes.}}
The deployment metrics separate transfer, implicit invocation, shortcut resistance, and composition. 
This separation matters because failures appear on different axes. 
GPT-5.4 under \textsc{Curated-Start} \cond{+ Always} improves ESR, ARSR, and CompSR, but decreases CSSR, suggesting an implicit-invocation failure when the skill need is embedded in a broader context. 
Gemini 3 Flash under the same condition gains strongly on CSSR and improves CompSR, yet drops sharply on ARSR, showing shortcut vulnerability. 
GPT-5.4 under \textsc{Self-Generated} \cond{Experience} improves LSR, RSR, and ARSR, but loses on CompSR, indicating weak modularity. 
Thus, a single success rate would hide whether a skill fails through over-specialized patching, missed invocation, shortcut reliance, or weak composition.

\textbf{\noindent{Raw trajectories reveal a lossy abstraction bottleneck.}}
\Cref{fig:rawtraj_vs_skills_delta} provides the clearest evidence that abstraction from episodes into skills remains unreliable. 
If skill abstraction preserved the reusable procedure, skill-based conditions should match or exceed \textsc{Raw-Trajectory}. 
Instead, the heatmap is predominantly negative across models, variants, and metrics. 
For example, GPT-5.4 under \textsc{Self-Generated} \cond{Experience} exceeds \textsc{Raw-Trajectory} on LSR and RSR, but falls substantially on ESR, CSSR, ARSR, and CompSR. 
This suggests that distilled skills may recover behavior near the original episode while losing contextual and procedural cues needed for transfer, robustness, and composition. 

\textbf{\noindent{More skill updates are not monotonically better.}}
The \cond{+ Always} policy reveals a coverage--drift trade-off. 
In the \textsc{Self-Generated} setting, more frequent updates can improve deployment-side coverage: GPT-5.2-Codex improves from \cond{Experience} to \cond{+ Always} on ESR and CSSR, and Gemini 2.5 Pro also gains on several deployment metrics. 
However, always updating can reduce local gains, as Gemini 3.1 Pro drops from strong LSR and RSR improvements under \cond{Experience} to much smaller gains under \cond{+ Always.} 
In the \textsc{Curated-Start} setting, \cond{+ Always} also does not dominate \cond{+ Revision.} Thus, episodic experience does not become a reusable skill simply because it is written into the library more often. 
Agents need update policies that preserve generality, filter episode-specific detail before it persists, and retain the procedural cues needed for future deployment under context shift, adversarial shortcuts, and composition.

\subsection{Capacity Diagnostic: Does More Skill Capacity Help?}
\label{sec:capacity_diagnostic}

\begin{figure*}[t]
\centering
\includegraphics[width=\textwidth]{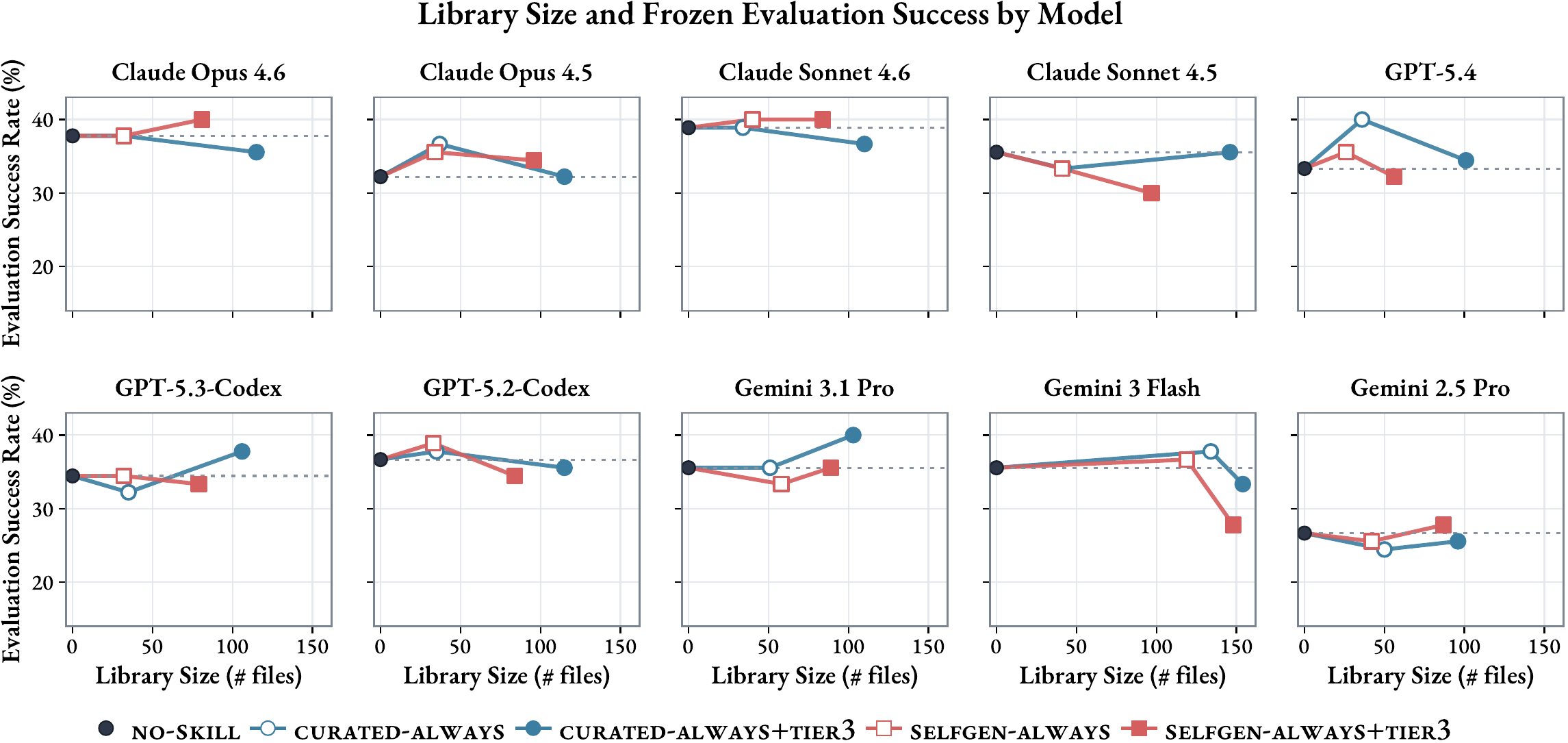}
\caption{
\textbf{Library size versus frozen evaluation success.}
Each panel corresponds to one model. The horizontal axis is the final library
size, measured as the number of \texttt{SKILL.md}, \texttt{references/},
\texttt{scripts/}, and \texttt{assets/} files. The vertical axis reports ESR.
Open markers denote ordinary \textsc{Always} revision; filled markers denote
\textsc{Always+Tier3}. Lines connect each memory condition to the same model's
\textsc{No-Skill} run.
}
\label{fig:library_size_vs_esr_model_specific}
\end{figure*}

The \textsc{Raw-Trajectory} comparison suggests that compact skill abstraction may discard useful procedural evidence. 
We therefore ask whether agents can extract enough reusable information from execution traces to enrich the skill library beyond a single \texttt{SKILL.md} file. 
These traces contain concrete evidence, including repeatedly re-derived commands, validation scripts the agent wished it had, schema details discovered at runtime, lookup tables, examples, and templates. 
A capable Skill Author should be able to turn such evidence into persistent resources under \texttt{scripts/}, \texttt{references/}, and \texttt{assets/}. 
The question is therefore not simply whether larger libraries are better, but whether additional library capacity helps preserve reusable procedural structure in a form that future agents can load and apply.

We use Tier-3 forcing as a diagnostic for this question. 
In the \cond{Always+Tier3} ablation, each eligible revision must include at least one new or updated resource under an existing skill's \texttt{scripts/}, \texttt{references/}, or \texttt{assets/} directory. 
The non-Tier-3 \cond{Always} condition already allows the Skill Author to add such resources when they materially help, whereas the Tier-3 setting makes resource bundling mandatory. 
This yields a controlled comparison between ordinary free-form skill editing and aggressive resource bundling while keeping the trajectory evidence, verifier feedback, existing skill state, and Skill Author call surface unchanged across conditions. 
The exact Tier-3 authoring instruction and parser constraint are provided in Appendix~\Cref{app:tier3_prompt_comparison}, including the distinction between optional resource use and mandatory Tier-3 file creation. 
If capacity is the main bottleneck, then \cond{Always+Tier3} should produce richer libraries while preserving or improving frozen deployment success. 
If the real problem is selective abstraction rather than storage, then the library may grow even as ESR stays flat or declines across harder deployment roles.

\Cref{fig:library_size_vs_esr_model_specific} separates capacity expansion from functional improvement. 
Forced Tier-3 authoring does make agents write richer libraries: the \cond{Always+Tier3} variants generally add more files under \texttt{scripts/}, \texttt{references/}, and \texttt{assets/}. 
This indicates that agents can identify trace-derived material that appears reusable. 
However, larger libraries do not reliably improve frozen deployment. 
Many points in \Cref{fig:library_size_vs_esr_model_specific} move rightward without moving upward: the number of persisted files increases, but ESR stays flat or decreases. 
Full per-model and per-metric Tier-3 ablation results are reported in Appendix~\Cref{app:tier3_full_results}; they show the same pattern at the metric level, where additional resources sometimes improve acquisition or replay metrics but do not consistently improve frozen deployment, context-shift invocation, adversarial robustness, or multi-skill composition.

The positive cases show that additional resources can help when they capture stable procedures that the task-solving agent can actually reuse. 
Claude Opus 4.6 improves under \textsc{SelfGen-Always+Tier3}, from 37.8\% to 40.0\% ESR, and Gemini 3.1 Pro improves under \textsc{Curated-Always+Tier3}, from 35.6\% to 40.0\%. 
These results suggest that some procedural details are not fully preserved in the compact \texttt{SKILL.md} body alone. 
In such cases, Tier-3 files can serve their intended role by preserving reusable validation routines, reference material, templates, or executable helpers that future agents can load on demand.
\begin{figure*}[t]
    \centering
    \includegraphics[width=0.83\linewidth]{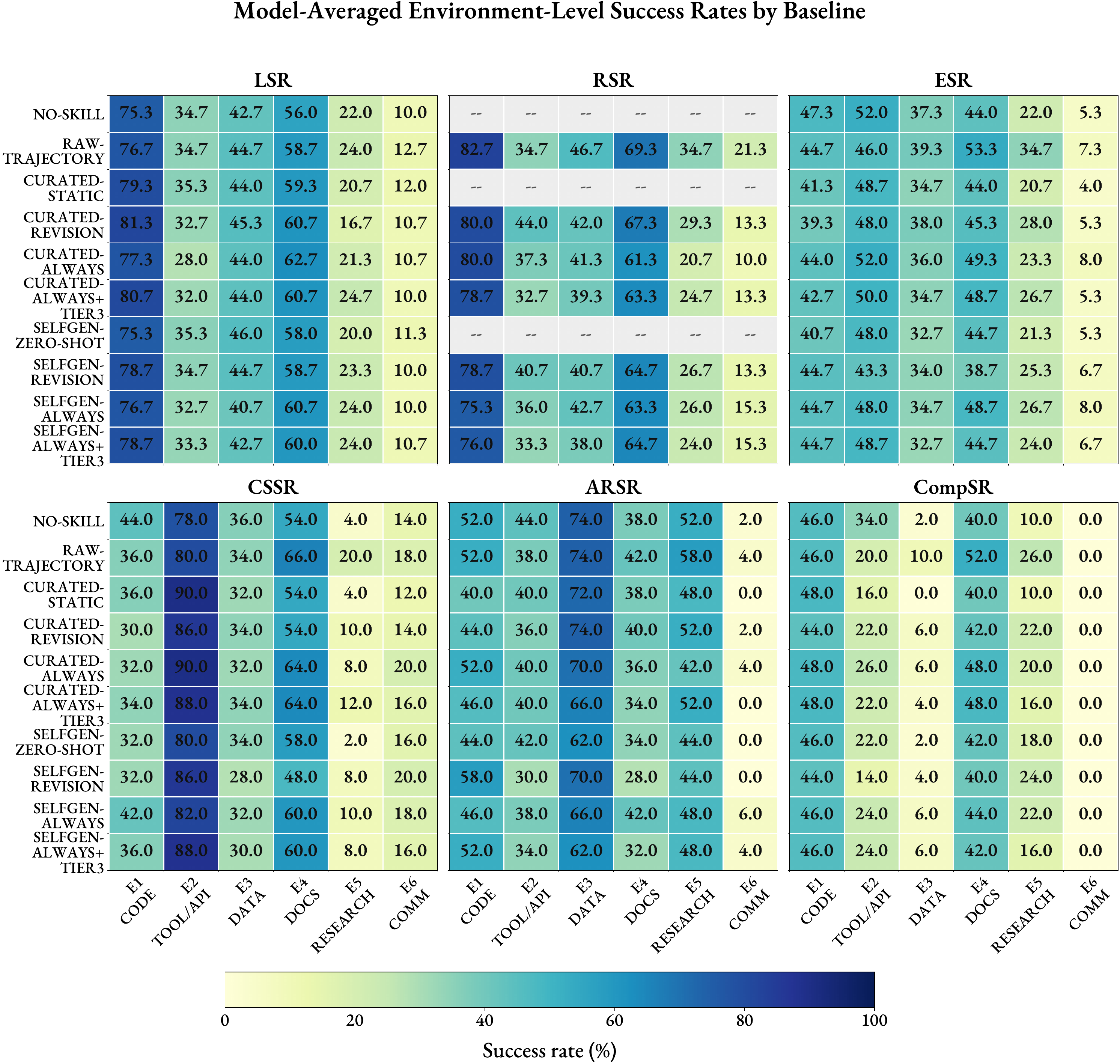}
    \caption{
    \textbf{Environment-level success rates by baseline.}
    Each cell reports success averaged over model--harness runs for one baseline and environment.
    Columns denote E1 Code Debugging \& Modification, E2 Tool \& API Orchestration, E3 Data Processing \& Structured Query, E4 Document Parsing \& Transformation, E5 Research \& Information Synthesis, and E6 Communication \& Scheduling.
    LSR and RSR measure acquisition and replay success; ESR measures frozen deployment success.
    CSSR, ARSR, and CompSR decompose deployment into context-shift, adversarial, and composition roles.
    Gray cells indicate variants without replay results.
    }
    \label{fig:environment_metric_heatmaps}
    \vspace{-1mm}
\end{figure*}

The failure cases show where capacity expansion breaks down. 
GPT-5.4 reaches its highest ESR with \textsc{Curated-Always}, while the larger \textsc{Curated-Always+Tier3} and \textsc{SelfGen-Always+Tier3} libraries reduce ESR. 
Gemini 3 Flash is the clearest overload case: its \textsc{SelfGen-Always+Tier3} library is among the largest, but ESR drops from 35.6\% under \textsc{No-Skill} to 27.8\%. 
Claude Sonnet 4.5 likewise drops from 35.6\% to 30.0\% under \textsc{SelfGen-Always+Tier3}. These declines suggest that forced resource bundling can preserve details that appear useful locally but do not transfer beyond the original acquisition context. 
The added resources may encode episode-specific assumptions, over-specific validators, stale context, or weakly triggered files that increase retrieval burden and distract the agent during frozen evaluation.

Overall, the answer to the capacity question is only partially positive. 
Agents can write more files and package more material as persistent skill resources, but they do not yet reliably preserve the right information. 
The limiting factor is selective enrichment: deciding which details are stable enough to persist, how to organize them as procedural resources, and when a future agent should load them. 
Tier-3 therefore exposes the same abstraction bottleneck from another angle. 
More capacity helps only when it is controlled; otherwise, richer libraries become procedural clutter rather than more reliable skills.

\subsection{Environment-Level Success Patterns}
\label{sec:environment_success_patterns}

\Cref{fig:environment_metric_heatmaps} shows that environment structure is another large source of variation. 
Across baselines, the gap between the easiest and hardest environments is 67.3 percentage points for LSR and 42.1 points for ESR. 
By contrast, the corresponding ranges across baselines are only 1.9 and 5.4 points. 
This suggests that \skillevolbench{} is not merely measuring a single global effect of adding memory. 
Instead, it exposes domain-specific differences in how easily episodic experience can be converted into reusable procedural knowledge.

The deployment roles reveal distinct bottlenecks. 
E2 Tool/API has the highest context-shift success, with mean CSSR of 84.7\%, indicating that agents can often reuse procedural patterns for changed API contexts once the relevant interface discipline is available. 
E3 Data has the highest adversarial success, with mean ARSR of 69.8\%, but its mean CompSR is only 4.5\%. 
Thus, robustness to adversarial perturbations and skill composition are not interchangeable deployment capabilities. 
Similarly, E1 Code and E4 Docs are comparatively strong on composition, while E6 Communication remains difficult across all conditions. 
In particular, E6 obtains 0.0\% CompSR for every baseline, highlighting communication and scheduling as a hard setting for multi-constraint procedural reuse.

The heatmaps also show that skill-based abstraction does not uniformly dominate direct episodic reuse. 
\textsc{Raw-Trajectory} is the strongest baseline by mean RSR, ESR, ARSR, and CompSR, with averages of 48.2\%, 37.6\%, 44.7\%, and 25.7\%, respectively. 
It is especially competitive in E4 Document Parsing \& Transformation and E5 Research \& Information Synthesis, where retaining concrete traces can preserve useful task-specific details that may be lost during abstraction. 
At the same time, replay improvements in structured environments such as E1 Code and E4 Docs indicate that experience can become reusable after acquisition. 
Together, these patterns suggest that skill evolution is most reliable when the environment supports stable procedural abstractions, whereas open-ended information and communication tasks remain sensitive to missing context, underspecified constraints, and brittle composition.

\subsection{Cost--Success Analysis}
\label{sec:cost_success_analysis}

\begin{figure*}[t]
\centering
\includegraphics[width=\linewidth]{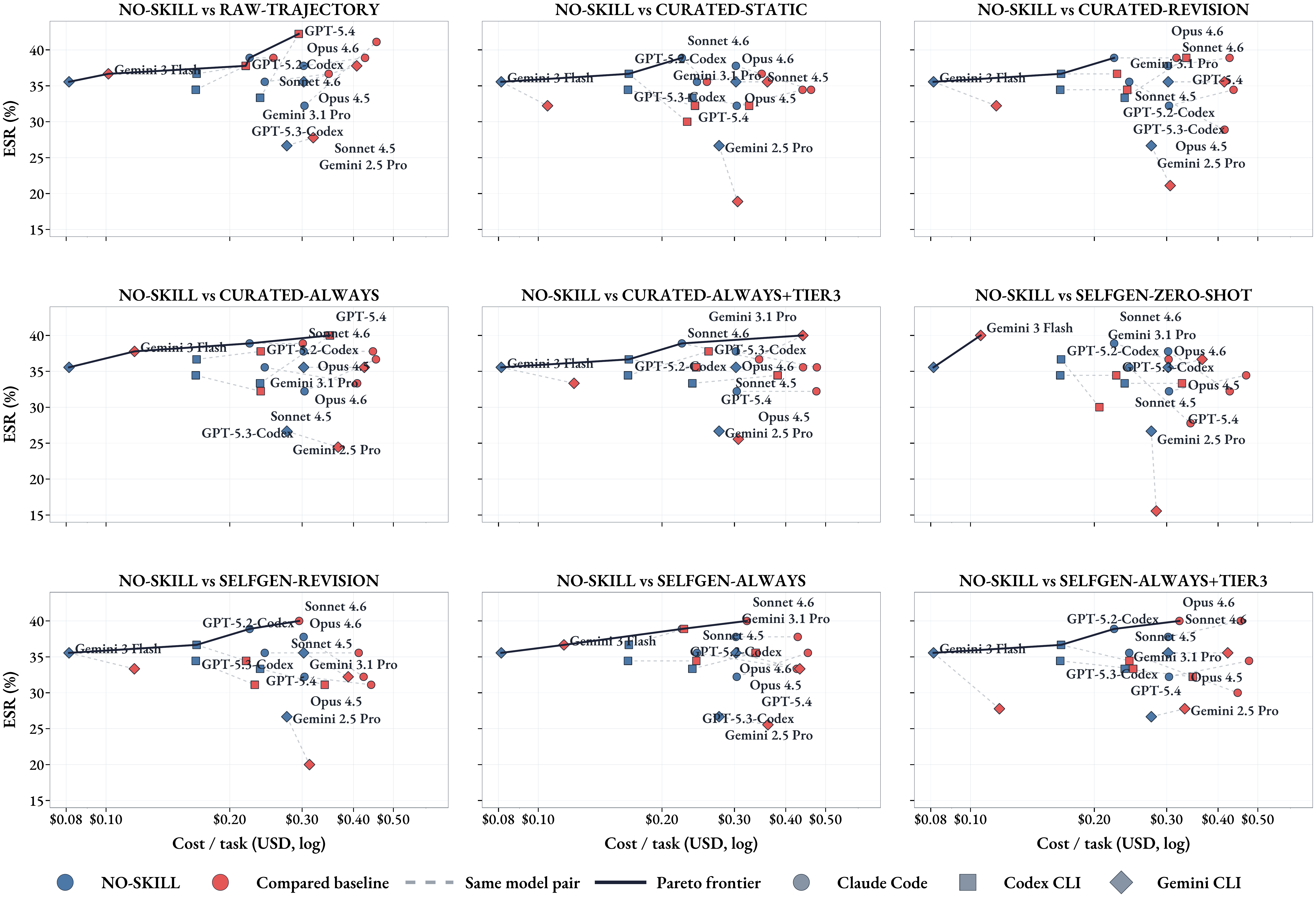}
\caption{
\textbf{Model-specific cost--success trade-offs relative to \textsc{No-Skill}.}
Each panel compares \textsc{No-Skill} against one memory variant using the same ten models.
The \(x\)-axis reports cost per attempted task in USD on a log scale, and the \(y\)-axis reports frozen evaluation success rate (ESR).
Blue markers denote the \textsc{No-Skill} run and red markers denote the compared variant.
Dashed gray segments connect the two points from the same model.
Marker shape indicates the agent harness.
The dark line marks the Pareto frontier within each pairwise comparison.
}
\label{fig:noskill_pairwise_cost_success}
\end{figure*}

\Cref{fig:noskill_pairwise_cost_success} shows that memory mechanisms do not form a uniform cost--success frontier across models. 
Most variants increase cost per attempted task relative to \textsc{No-Skill}, but the corresponding ESR gain is highly model- and variant-dependent. 
This is visible from the direction of the paired segments: many move rightward with little vertical movement, while only a smaller subset move both rightward and upward.

\textbf{\noindent{Static curated skills are not sufficient.}}
\textsc{Curated-Static} underperforms \textsc{No-Skill} on average. 
It improves 2 models, ties 1, and hurts 7, with a mean ESR change of -2.44 percentage points. 
This suggests that fixed curated procedures are not automatically useful under deployment shift. 
When the skill cannot be revised, it may overfit to the canonical procedure or introduce misleading constraints for later roles. 
The cost also increases by \$0.077 per attempted task on average, so the static skill condition is generally dominated by \textsc{No-Skill} and \textsc{Raw-Trajectory}.

\textbf{\noindent{Revision helps curated skills, but only partially.}}
Allowing curated skills to be revised makes the curated family more competitive, but the effect remains mixed. 
\textsc{Curated-Revision} has an average change of -0.67 percentage points, with 3 wins, 4 ties, and 3 losses. 
\textsc{Curated-Always} is the strongest curated variant, with a mean gain of +0.78 percentage points and 4 wins, 3 ties, and 3 losses. 
The benefit is concentrated in particular models: GPT-5.4 improves by +6.7 percentage points under \textsc{Curated-Always}, and Opus 4.5 improves by +4.4 percentage points. 
Thus, frequent curated-skill updating can help, but it does not produce a model-agnostic advantage across the full set of deployment conditions.

\textbf{\noindent{Tier-3 authoring does not reliably improve the frontier.}}
\textsc{Curated-Tier3} and \textsc{SelfGen-Tier3} do not produce stable gains despite their higher authoring cost. 
\textsc{Curated-Tier3} has a mean ESR change of 0.00 percentage points, with 3 wins, 2 ties, and 5 losses, while increasing cost by \$0.119 per attempted task on average. 
\textsc{SelfGen-Tier3} averages -1.11 percentage points, with 4 wins, 1 tie, and 5 losses. 
These results suggest that a more detailed or more strongly constrained authoring process is not automatically better. 
In this setting, additional authoring effort can increase cost without reliably improving deployment success.

\textbf{\noindent{Self-generated skills need frequent updates.}}
The self-generated variants show a clear distinction between update policies. 
\textsc{SelfGen-Revision}, which updates only after eligible failures, is weak: it improves only 1 model, ties 1, and hurts 8, with an average change of -2.56 percentage points. 
\textsc{SelfGen-Always} is substantially healthier, improving 5 models, tying 2, and hurting 3, with an average change of +0.44 percentage points. 
This supports the interpretation that self-generated procedural memory benefits from dense update opportunities. 
Failure-only revision appears too sparse or too reactive to reliably produce reusable skills.

\textbf{\noindent{Zero-shot skill generation is brittle.}}
\textsc{SelfGen-ZeroShot} is also negative on average, with a mean ESR change of -2.56 percentage points. 
It helps Gemini 3 Flash by +4.4 percentage points, but hurts Gemini 2.5 Pro by -11.1 percentage points. 
The large spread indicates that metadata-only skills can act as useful prior knowledge for some models but harmful procedural bias for others. 
This brittleness motivates separating zero-shot initialization from experience-based skill evolution in the main analysis.

\textbf{\noindent{Model dependence.}}
The paired plots also reveal strong model dependence. 
Opus 4.5 benefits from 7 of the 9 compared variants, with an average ESR change of +2.72 percentage points. 
GPT-5.4 benefits from 5 variants, averaging +2.22 percentage points. 
In contrast, Gemini 2.5 Pro is harmed by 7 variants and averages -3.70 percentage points. 
Thus, the usefulness of procedural memory is not only a property of the memory mechanism; it also depends on the base model's ability to interpret, select, and apply the provided memory.

\textbf{\noindent{Takeaway.}}
The cost--success analysis reinforces three main conclusions. 
First, skill abstraction is not automatically beneficial: static curated skills and sparse self-generated revision often add cost without improving ESR. 
Second, the most promising skill-based conditions are the always-update variants, but their gains remain model-specific and do not yet dominate direct trajectory retrieval. 
Third, Tier-3 resource bundling shows that larger and more expensive libraries can fail when capacity expansion is not matched by selective procedural abstraction.
\section{Conclusion}
\label{sec:conclusion}

We introduced \skillevolbench{}, a diagnostic benchmark for evaluating whether agents can transform episodic task experience into reusable procedural skills. 
Rather than only measuring whether skills help at inference time, \skillevolbench{} targets the missing step from experience reuse to skill formation. 
Its role-conditioned task families, verifier-backed feedback, frozen deployment phase, replay setting, and \textsc{Raw-Trajectory} control help separate local task recovery from transferable procedural reuse. 

Across ten model configurations and three agent harnesses, our experiments show that current agentic LLMs often adapt locally but rarely form reliable reusable skills. 
Skill-based conditions can improve acquisition or replay, yet their gains are unstable under context shift, adversarial shortcuts, and composition. 
The comparison with \textsc{Raw-Trajectory} reveals a lossy abstraction bottleneck: agents often use episodic traces more effectively than the distilled skills derived from them. 
The Tier-3 capacity diagnostic further shows that writing larger skill libraries is not sufficient; additional resources help only when they preserve stable procedures rather than episode-specific clutter. 
Environment-level and cost--success analyses show that skill evolution remains strongly shaped by task structure, deployment role, authoring cost, and base-model ability. 

Together, these results suggest that the key challenge is not simply storing more experience, revising skills more often, or giving agents larger resource libraries. 
The core problem is selective procedural abstraction: agents must preserve the details that support future invocation, verification, robustness, and composition while filtering out local repairs and episode-specific noise. 
\skillevolbench{} provides a testbed for measuring progress on this step from one-off task experience to durable procedural knowledge under deployment shift.

\bibliography{reference}
\bibliographystyle{unsrtnat}

\newpage

\appendix
\section{Complete Family Catalog}
\label{app:family_catalog}

\newlength{\familytabwidth}
\newcommand{\setfamilytabwidth}{%
  \setlength{\familytabwidth}{\dimexpr\linewidth-6\tabcolsep\relax}%
}

\newcommand{\familycatalogheader}{%
\toprule
\multicolumn{1}{>{\raggedright\arraybackslash}p{0.08\familytabwidth}}{\textbf{Family}} &
\multicolumn{1}{>{\raggedright\arraybackslash}p{0.18\familytabwidth}}{\textbf{Skill}} &
\multicolumn{1}{>{\raggedright\arraybackslash}p{0.25\familytabwidth}}{\textbf{Environment}} &
\multicolumn{1}{>{\raggedright\arraybackslash}p{0.49\familytabwidth}}{\textbf{Family Description}} \\
\midrule
}

This appendix catalogs the full set of environment-level skill families in SkillEvolBench. Each family corresponds to one procedural skill and contains six role-instantiated tasks.

\begingroup
\small
\setlength{\tabcolsep}{1.5pt}
\renewcommand{\arraystretch}{1.05}
\setfamilytabwidth
\begin{longtable}{@{}%
>{\raggedright\arraybackslash}p{0.08\familytabwidth}%
>{\raggedright\arraybackslash}p{0.18\familytabwidth}%
>{\raggedright\arraybackslash}p{0.25\familytabwidth}%
>{\raggedright\arraybackslash}p{0.49\familytabwidth}%
@{}}
\caption{Complete skill-family catalog. Descriptions are taken from the benchmark family metadata.}\label{tab:family_catalog_no_exposure}\\
\familycatalogheader
\endfirsthead
\familycatalogheader
\endhead

E1-LS1 & systematic-error-diagnosis & E1: Code Debugging Modification & Systematically diagnose and fix runtime errors in multi-file Python and JavaScript/TypeScript projects. Use this skill when a task includes a traceback, stack trace, failing command, failing test, or clear crash across 3-7 source files and you need to trace the root cause through the call chain. Focus on failures with explicit error output or crashes. Trigger on requests like "fix the bug", "resolve the traceback", "debug this project", or "make the failing test pass" when the failure already points to a concrete runtime error. \\
\midrule
E1-LS2 & dependency-conflict-resolution & E1: Code Debugging Modification & Diagnose and resolve Python package version conflicts in pip-based projects. Use this skill when `pip install` fails with resolver errors, when `pip check` reports incompatible versions, or when a Python project's dependency graph contains conflicting version constraints that must be reconciled in `requirements.txt` or `pyproject.toml`. Focus on single-project pip workflows with explicit version conflicts and shared dependencies. \\
\midrule
E1-LS3 & safe-refactoring & E1: Code Debugging Modification & Safely restructure Python code to improve clarity, reduce duplication, and increase maintainability without changing external behavior. Use this skill when the user asks to refactor, clean up, simplify, rename, extract functions or classes, remove dead code, or reorganize Python code while keeping the existing behavior intact. Focus on Python projects with an existing test suite and structural changes only. \\
\midrule
E1-LS4 & multi-file-bug-fix & E1: Code Debugging Modification & Trace and fix bugs that span multiple files in Python and JavaScript projects. Use this skill when the visible symptom appears in one file but the root cause is likely upstream or downstream in another file, and the fix requires coordinated updates across several files. Trigger when a user reports wrong output, a broken flow, a changed function contract, a module move, or any bug where one local edit is unlikely to be enough. Focus on projects with roughly 4-7 files involved in the debugging path. \\
\midrule
E1-LS5 & merge-conflict-resolution & E1: Code Debugging Modification & Resolve git merge conflicts in Python and JavaScript code by combining changes from two branches correctly. Use this skill when `git merge` reports conflicted files, when a file contains conflict markers like `<<<<<<<`, `=======`, and `>>>>>>>`, or when the user asks for help merging two branches without discarding either side's important changes. Focus on conflicts that git reports explicitly in source files with conflict markers. \\
\midrule
E2-LS1 & pre-call-parameter-validation & E2: Multi Step Tool Api Orchestration & Validate REST API request parameters locally before sending the request. Use this skill when parameters come from user input, config, or upstream code and an invalid request would waste quota or fail predictably. Focus on single-request validation for required fields, types, ranges, enums, and string formats before making the API call. \\
\midrule
E2-LS2 & retry-and-backoff & E2: Multi Step Tool Api Orchestration & Retry synchronous REST API calls when server errors occur. Use this skill when an HTTP request may temporarily fail with a 5xx response and the caller needs a simple retry loop with a fixed delay between attempts. Focus on straightforward retry handling for synchronous API calls with status-code-based decision rules. \\
\midrule
E2-LS3 & pagination-complete-retrieval & E2: Multi Step Tool Api Orchestration & Retrieve the complete dataset from offset-based paginated REST APIs by fetching every page in sequence. Use this skill when an API returns results in numbered pages such as `page=1\&per\_page=25` and the user needs all records, not just the first page. Focus on page-number plus page-size pagination with JSON results that can be combined into one list. \\
\midrule
E2-LS4 & multi-step-orchestration & E2: Multi Step Tool Api Orchestration & Chain two API calls into a simple workflow where the output of the first call is used as input for the second call. Use this skill when a request pattern looks like authenticate then fetch, look up an ID then retrieve details, or any other two-step sequence where Call B depends directly on a value extracted from Call A. Focus on simple two-step API chains only. \\
\midrule
E2-LS5 & response-validation-fallback & E2: Multi Step Tool Api Orchestration & Validate REST API responses before using the data, and fall back to a secondary source when validation fails. Use this skill when an HTTP response must pass top-level checks for status code, content type, JSON parsing, and required top-level fields before the payload is trusted. Focus on simple response validation plus optional fallback handling. \\
\midrule
E3-LS1 & schema-inspection-before-query & E3: Data Processing Structured Query & Inspect CSV files and SQL tables before writing queries or data-processing code. Use this skill whenever the user asks for filtering, aggregation, joins, or analysis on tabular data and the schema must be checked first. Focus on quick practical inspection of columns, types, shape, sample rows, and join-key presence before any query is written. \\
\midrule
E3-LS2 & type-normalization-before-sort & E3: Data Processing Structured Query & Normalize inconsistent pandas column types before sorting or comparing values. Use this skill when a dataframe column mixes date formats, stores numbers as strings, or represents booleans in multiple textual forms and must be standardized before order-dependent operations. Focus on date, numeric, and boolean normalization in pandas DataFrames. \\
\midrule
E3-LS3 & key-alignment-before-merge & E3: Data Processing Structured Query & Align join keys across data sources before merging. Use this skill when two tables or DataFrames need to be joined but the key columns may differ in name, dtype, or string casing. Focus on simple key alignment before merge - column name matching, type matching, case normalization, and basic merge-result validation. \\
\midrule
E3-LS4 & null-safe-filtering-aggregation & E3: Data Processing Structured Query & Handle missing values correctly during data filtering and aggregation in pandas. Use this skill when a DataFrame contains NaN values and the task involves filtering rows, filling missing values, or computing aggregates such as mean, sum, or count. Focus on pandas NaN handling with isna(), dropna(), fillna(), and built-in aggregate functions before drawing conclusions from filtered or aggregated data. \\
\midrule
E3-LS5 & result-sanity-reconciliation & E3: Data Processing Structured Query & Verify processed tabular results with basic sanity checks before using or returning them. Use this skill after a pandas or SQL data-processing pipeline when you need to confirm that the final output has a reasonable row count, contains the expected columns, looks plausible in sample rows, and does not contain unexpected NaN values in critical fields. Focus on simple final-output verification with df.shape, df.columns, df.describe(), and df.isna().sum(). \\
\midrule
E4-LS1 & structured-field-extraction & E4: Document Parsing Extraction Transformation & Extract structured data from semi-structured documents such as invoices, forms, and reports, then convert the extracted fields to JSON. Use this skill when a document contains label-value pairs, markdown-style tables, or section headers that organize fields into simple categories. Focus on flat key-value extraction from single-format documents, with basic field validation after extraction. \\
\midrule
E4-LS2 & cross-format-migration & E4: Document Parsing Extraction Transformation & Convert flat documents from one format to another while preserving source content. Use this skill when the task is to migrate content such as Markdown, CSV, or simple key-value text into JSON, and the goal is to keep all body content represented in the target format. Focus on flat-format migration with explicit source-to-target mapping, parser-based conversion, and a final source-versus-target verification step. \\
\midrule
E4-LS3 & template-fill-from-context & E4: Document Parsing Extraction Transformation & Fill a template document using information found in one or two context documents. Use this skill when a template contains blank fields or placeholders and the task is to locate matching information in surrounding context, insert it into the template, and return a completed version. Focus on straightforward fill-from-context work with one template, one or two context documents, simple field-name matching, and basic output review. \\
\midrule
E4-LS4 & document-diff-comparison & E4: Document Parsing Extraction Transformation & Compare two versions of a plain text or markdown document and produce a clear change report. Use this skill when the task is to load two document versions, identify additions, deletions, and modifications, and report each change with its location, original text, and new text. Focus on text-level diff comparison for plain text and markdown documents using basic line-by-line or paragraph-by-paragraph comparison. \\
\midrule
E4-LS5 & multi-source-merge-reconciliation & E4: Document Parsing Extraction Transformation & Merge flat data from 2-3 document sources into a single master document using exact key matching. Use this skill when records from multiple sources need to be combined by a shared identifier, duplicate entities need to be unified, conflicting field values need to be flagged, and a source-priority rule should decide which value is kept in the merged output. Focus on exact-key matching across 2-3 sources with flat data structures. \\
\midrule
E5-LS1 & multi-source-search-filter & E5: Research Information Synthesis & Search and filter the most relevant documents from a collection of 10-20 sources on a given topic. Use this skill when the task is to quickly scan a set of articles, papers, reports, or similar documents, score them for topical relevance, select the top 5 candidates, and rank those selected sources from most to least relevant. Focus on relevance-based selection using keyword matching, topic fit, specific data/examples, and source credibility. \\
\midrule
E5-LS2 & evidence-grounded-comparison & E5: Research Information Synthesis & Compare 3-5 options using evidence found in provided source documents. Use this skill when the task is to define comparison dimensions, search source material for evidence on each option, build a comparison matrix with source citations, identify which option is strongest on each dimension, and write a summary recommendation based on the completed matrix. Focus on evidence-based comparison tables built from provided documents. \\
\midrule
E5-LS3 & citation-verification & E5: Research Information Synthesis & Verify that citations in a document accurately represent their sources. Use this skill when a paper, review article, report, or essay makes claims about external sources and the task is to check whether those cited sources actually support the attributed statements. Focus on verifying that claims, quotes, and numerical assertions are supported by the cited source content. \\
\midrule
E5-LS4 & constrained-summarization & E5: Research Information Synthesis & Generate summaries that must satisfy explicit constraints such as a word limit, required key points, and a requested presentation format. Use this skill when the task is to condense a long document into a shorter summary while staying within a stated word cap, ensuring required points are included, and checking the final draft against those constraints before returning it. Focus on word-limited summaries with required key point coverage. \\
\midrule
E5-LS5 & contradiction-detection & E5: Research Information Synthesis & Find direct factual contradictions between multiple information sources. Use this skill when two or more documents make claims about the same facts and the task is to extract those claims, group them by topic, compare the stated values, and report any contradictions with their sources. Focus on direct factual contradictions such as different numbers, dates, names, or statistics for the same fact. \\
\midrule
E6-LS1 & inbox-triage-prioritization & E6: Communication Scheduling Operations & Classify and prioritize incoming emails using simple category labels and explicit urgency markers. Use this skill when the task is to scan a batch of emails, assign each one to a fixed category, detect clear urgency keywords in the subject or opening sentences, and output a sorted list with High priority first, then Medium, then Low. Focus on keyword-based classification and explicit urgency markers. \\
\midrule
E6-LS2 & context-aware-reply-drafting & E6: Communication Scheduling Operations & Draft professional reply emails that reference the prior context of a business email thread. Use this skill when a user provides a thread of 3-5 messages and wants a reply that reads the conversation history, identifies the latest question or request, answers it directly, references relevant earlier context when helpful, and matches the tone of the original sender. Focus on concise professional reply drafting for standard business email threads. \\
\midrule
E6-LS3 & follow-up-tracking & E6: Communication Scheduling Operations & Identify email threads that are overdue for a response by checking the direction of the last message, whether it contains a question or request, and how many business days have passed since it was received. Use this skill when the task is to scan one or more email threads, find items waiting on your reply, and produce an overdue follow-up list sorted by age. Focus on basic overdue detection based on last-message date and direction. \\
\midrule
E6-LS4 & meeting-scheduling-constraints & E6: Communication Scheduling Operations & Find a common meeting time for 2-3 participants across 2 time zones using fixed UTC offsets. Use this skill when the task is to list participants and time zones, convert business-hour availability into UTC, find the overlapping UTC range, convert that overlap back into local times, and suggest a specific meeting time within the overlap. Focus on scheduling with fixed UTC offsets and standard business hours. \\
\midrule
E6-LS5 & thread-action-extraction & E6: Communication Scheduling Operations & Extract explicit action items from formal business email threads. Use this skill when the task is to read an email conversation, identify direct requests, named assignments, and explicit sender commitments, and output structured action items with a task description, assignee, deadline, and source email. Focus on extracting explicit action items from professional email threads. \\
\bottomrule
\end{longtable}
\endgroup
\section{Task Design Catalog}
\label{app:task_design_catalog}

\newlength{\tasktabwidth}
\newcommand{\settasktabwidth}{%
  \setlength{\tasktabwidth}{\dimexpr\linewidth-8\tabcolsep\relax}%
}

\newcommand{\taskcatalogheader}{%
\toprule
\multicolumn{1}{>{\centering\arraybackslash}p{0.08\tasktabwidth}}{\textbf{Task}} &
\multicolumn{1}{>{\centering\arraybackslash}p{0.08\tasktabwidth}}{\textbf{Family}} &
\multicolumn{1}{>{\centering\arraybackslash}p{0.12\tasktabwidth}}{\textbf{Role / Phase}} &
\multicolumn{1}{>{\centering\arraybackslash}p{0.33\tasktabwidth}}{\textbf{Title}} &
\multicolumn{1}{>{\centering\arraybackslash}p{0.39\tasktabwidth}}{\textbf{Design Signals}} \\
\midrule
}

\noindent This appendix lists all task instances. Design signals record gap exposure, trap annotations, and composition requirements from benchmark metadata; task instruction bodies are omitted.

\subsection{E1: Code Debugging \& Modification}
\begingroup
\scriptsize
\setlength{\tabcolsep}{1.5pt}
\renewcommand{\arraystretch}{1.05}
\settasktabwidth
\begin{longtable}{@{}%
>{\raggedright\arraybackslash}p{0.08\tasktabwidth}%
>{\raggedright\arraybackslash}p{0.08\tasktabwidth}%
>{\raggedright\arraybackslash}p{0.12\tasktabwidth}%
>{\raggedright\arraybackslash}p{0.33\tasktabwidth}%
>{\raggedright\arraybackslash}p{0.39\tasktabwidth}%
@{}}
\caption{Task design catalog for E1: Code Debugging \& Modification.}\label{tab:task_catalog_e1}\\
\taskcatalogheader
\endfirsthead
\taskcatalogheader
\endhead
E1-LS1-T1 & E1-LS1 & canonical / learning & Fix the intermittent 500s in our analytics API & gap=none \\
E1-LS1-T2 & E1-LS1 & enriched / learning & Close the auth bypass in the FinGuard API & gap=gap1-no-crash-behavior-error \\
E1-LS1-T3 & E1-LS1 & variant / learning & Fix the report job so timeouts are not silently treated as real zeros & gap=gap2-async-silent-failure \\
E1-LS1-T4 & E1-LS1 & context-shift / evaluation & Finish the webhook work in this repo & gap=implicit-precondition-debugging \\
E1-LS1-T5 & E1-LS1 & adversarial / evaluation & Fix the tax calculation bug without patching around individual states & gap=shortcut-trap; traps=shortcut, generic\_trap \\
E1-LS1-T6 & E1-LS1 & composition / evaluation & Fix the blank profile page by tracing the full failure path & gap=multi-hop-error-cascade; composition=cross\_family; requires E1-LS1.systematic-error-diagnosis, E1-LS3.safe-refactoring, E1-LS4.multi-file-bug-fix \\
\midrule
E1-LS2-T1 & E1-LS2 & canonical / learning & Fix the runtime dependency issue in this Python project & gap=none \\
E1-LS2-T2 & E1-LS2 & enriched / learning & Fix the React 18 dependency conflict in this frontend app & gap=gap1-npm-peer-dependency \\
E1-LS2-T3 & E1-LS2 & variant / learning & Fix the Poetry dependency conflict in this monorepo & gap=gap2-monorepo-cross-package \\
E1-LS2-T4 & E1-LS2 & context-shift / evaluation & Finish the pending PR for CSV export & gap=implicit-blocker-in-context \\
E1-LS2-T5 & E1-LS2 & adversarial / evaluation & Fix the ML pipeline crash after the dependency update & gap=shortcut-pin-old-version-trap; traps=shortcut, generic\_trap \\
E1-LS2-T6 & E1-LS2 & composition / evaluation & Upgrade Alembic and make the SQLAlchemy stack compatible & gap=multi-layer-upgrade-migration; composition=cross\_family; requires E1-LS2.dependency-conflict-resolution, E1-LS4.multi-file-bug-fix \\
\midrule
E1-LS3-T1 & E1-LS3 & canonical / learning & Reduce duplicated validation in the intake handlers & gap=none \\
E1-LS3-T2 & E1-LS3 & enriched / learning & Replace the event switch with a strategy mapping without changing behavior & gap=gap1-fallthrough-sideeffects \\
E1-LS3-T3 & E1-LS3 & variant / learning & Refactor the processor hierarchy into composition without breaking TypeScript constraints & gap=gap2-typesystem-super-chain \\
E1-LS3-T4 & E1-LS3 & context-shift / evaluation & Add plugin support, but do it in a maintainable way & gap=implicit-refactor-before-feature \\
E1-LS3-T5 & E1-LS3 & adversarial / evaluation & Refactor the parse/transform/format chain without collapsing empty-input behavior & gap=gap1-boundary-behavior-preservation; traps=gap1-boundary-behavior-preservation \\
E1-LS3-T6 & E1-LS3 & composition / evaluation & Refactor the analysis function and use coverage to close the gaps & gap=coverage-driven-edge-case-diagnosis; composition=cross\_family; requires E1-LS3.safe-refactoring, E1-LS4.multi-file-bug-fix \\
\midrule
E1-LS4-T1 & E1-LS4 & canonical / learning & Fix the false-positive change detection in the ETL pipeline & gap=none \\
E1-LS4-T2 & E1-LS4 & enriched / learning & Fix the blank GraphQL profile fields & gap=gap1-cross-system-naming \\
E1-LS4-T3 & E1-LS4 & variant / learning & Fix the false mismatches between the two services & gap=gap2-json-precision-loss \\
E1-LS4-T4 & E1-LS4 & context-shift / evaluation & Run the migration and make sure the timeline stays correct & gap=implicit-timezone-bug-in-migration \\
E1-LS4-T5 & E1-LS4 & adversarial / evaluation & Fix the profile page error handling end to end & gap=frontend-workaround-trap; traps=generic\_trap \\
E1-LS4-T6 & E1-LS4 & composition / evaluation & Fix the order precision issue, then lock it down with tests and docs & gap=precision-fix-plus-tests-plus-docs; composition=cross\_family; requires E1-LS1.systematic-error-diagnosis, E1-LS4.multi-file-bug-fix \\
\midrule
E1-LS5-T1 & E1-LS5 & canonical / learning & Task: Resolve the Merge Before We Cut the Utility Rollout & gap=none \\
E1-LS5-T2 & E1-LS5 & enriched / learning & Task: Review a "Successful" Merge That Still Broke Validation & gap=gap1-semantic-conflict-after-clean-merge \\
E1-LS5-T3 & E1-LS5 & variant / learning & Task: Finish the Config Merge Without Breaking the YAML Structure & gap=gap2-format-sensitive-config-merge \\
E1-LS5-T4 & E1-LS5 & context-shift / evaluation & Task: Unblock the v2.1 Release Merge & gap=implicit-conflict-during-release-workflow \\
E1-LS5-T5 & E1-LS5 & adversarial / evaluation & Task: Resolve the Database Merge Without Dropping the Safer Query Changes & gap=shortcut-trap-accept-ours-drops-security-fix; traps=shortcut, generic\_trap \\
E1-LS5-T6 & E1-LS5 & composition / evaluation & Task: Finish the End-of-Sprint Feature Merge & gap=merge-plus-regression-diagnosis; composition=cross\_family; requires E1-LS1.systematic-error-diagnosis, E1-LS4.multi-file-bug-fix, E1-LS5.merge-conflict-resolution \\
\bottomrule
\end{longtable}
\endgroup

\subsection{E2: Tool \& API Orchestration}
\begingroup
\scriptsize
\setlength{\tabcolsep}{1.5pt}
\renewcommand{\arraystretch}{1.05}
\settasktabwidth
\begin{longtable}{@{}%
>{\raggedright\arraybackslash}p{0.08\tasktabwidth}%
>{\raggedright\arraybackslash}p{0.08\tasktabwidth}%
>{\raggedright\arraybackslash}p{0.12\tasktabwidth}%
>{\raggedright\arraybackslash}p{0.33\tasktabwidth}%
>{\raggedright\arraybackslash}p{0.39\tasktabwidth}%
@{}}
\caption{Task design catalog for E2: Tool \& API Orchestration.}\label{tab:task_catalog_e2}\\
\taskcatalogheader
\endfirsthead
\taskcatalogheader
\endhead
E2-LS1-T1 & E2-LS1 & canonical / learning & Task: Process the Weather Batch Without Sending Bad Requests & gap=none \\
E2-LS1-T2 & E2-LS1 & enriched / learning & Task: Validate Date Ranges Before Sending Search Requests & gap=gap1-cross-parameter-constraint-with-timezone-normalization \\
E2-LS1-T3 & E2-LS1 & variant / learning & Task: Import the Batch and Reject Only the Bad Rows & gap=gap2-batch-partial-reject-with-per-row-reporting \\
E2-LS1-T4 & E2-LS1 & context-shift / evaluation & Task: Turn the Sales Query Into a Valid Search Request & gap=implicit-validation-inside-search-pipeline \\
E2-LS1-T5 & E2-LS1 & adversarial / evaluation & Task: Stop Depending on Backend 400 Payloads & gap=remote-400-payload-trap; traps=generic\_trap \\
E2-LS1-T6 & E2-LS1 & composition / evaluation & Task: Complete the Validation and Enrichment Pipeline Before Calling Transactions & gap=validate-enrich-normalize-call-composition; composition=cross\_family; requires E2-LS1.pre-call-parameter-validation, E2-LS5.response-validation-fallback \\
\midrule
E2-LS2-T1 & E2-LS2 & canonical / learning & Task: Respect Retry-After When the API Rate Limits Us & gap=none \\
E2-LS2-T2 & E2-LS2 & enriched / learning & Task: Make the 503 Retry Strategy Production-Safe & gap=gap1-exponential-backoff-and-jitter-for-503 \\
E2-LS2-T3 & E2-LS2 & variant / learning & Task: Retry the Order Create Call Without Duplicating Orders & gap=gap2-non-idempotent-post-retry-requires-shared-idempotency-key \\
E2-LS2-T4 & E2-LS2 & context-shift / evaluation & Task: Retry Only the Failing Step in the Sync Pipeline & gap=implicit-precise-mid-pipeline-retry \\
E2-LS2-T5 & E2-LS2 & adversarial / evaluation & Task: Stop Blind Retries From Hiding Token Expiry & gap=trap-401-needs-refresh-not-blind-retry; traps=generic\_trap \\
E2-LS2-T6 & E2-LS2 & composition / evaluation & Task: Add a Real Circuit Breaker to the Downstream Client & gap=composition-retry-and-backoff-with-circuit-breaker-state-management; composition=cross\_family; requires E2-LS2.retry-and-backoff, E2-LS4.multi-step-orchestration \\
\midrule
E2-LS3-T1 & E2-LS3 & canonical / learning & Task: Retrieve the Full User Export from the Offset API & gap=none \\
E2-LS3-T2 & E2-LS3 & enriched / learning & Task: Fix the Event Stream Pagination Client & gap=gap1-cursor-based-pagination-with-variable-page-size \\
E2-LS3-T3 & E2-LS3 & variant / learning & Task: Make the Order Export Consistent While New Rows Arrive & gap=gap2-pagination-during-concurrent-inserts \\
E2-LS3-T4 & E2-LS3 & context-shift / evaluation & Task: Get the Full Catalog Export & gap=implicit-pagination-discovery \\
E2-LS3-T5 & E2-LS3 & adversarial / evaluation & Task: Retrieve the Full Set of Monthly Reports & gap=trap-trusting-first-total-pages-value; traps=generic\_trap \\
E2-LS3-T6 & E2-LS3 & composition / evaluation & Task: Merge Both Regional Order Feeds into One Complete Export & gap=composition-pagination-plus-retry-merge-sort; composition=cross\_family; requires E2-LS2.retry-and-backoff, E2-LS3.pagination-complete-retrieval \\
\midrule
E2-LS4-T1 & E2-LS4 & canonical / learning & Task: Authenticate, Collect Item Details, and Save the Full Export & gap=none \\
E2-LS4-T2 & E2-LS4 & enriched / learning & Task: Process Each User's Offers Through the Correct Branch & gap=gap1-conditional-branch-pipeline \\
E2-LS4-T3 & E2-LS4 & variant / learning & Task: Generate the Report Through the Async Job API & gap=gap2-async-polling-wait-pattern \\
E2-LS4-T4 & E2-LS4 & context-shift / evaluation & Task: Build the Cross-Regional Sales Summary & gap=implicit-multi-source-aggregation-flow \\
E2-LS4-T5 & E2-LS4 & adversarial / evaluation & Task: Fetch the Latest Dataset Through the Configured Endpoint & gap=trap-static-endpoint-assumption; traps=generic\_trap \\
E2-LS4-T6 & E2-LS4 & composition / evaluation & Task: Run the Full User Workflow End-to-End & gap=composition-orchestration-retry-validation; composition=cross\_family; requires E2-LS2.retry-and-backoff, E2-LS4.multi-step-orchestration, E2-LS5.response-validation-fallback \\
\midrule
E2-LS5-T1 & E2-LS5 & canonical / learning & Task: Product Feed Collection & gap=none \\
E2-LS5-T2 & E2-LS5 & enriched / learning & Task: Product sync during API rollout & gap=gap1-nested-wrapper-structural-change \\
E2-LS5-T3 & E2-LS5 & variant / learning & Task: Pricing feed validation & gap=gap2-semantic-anomaly \\
E2-LS5-T4 & E2-LS5 & context-shift / evaluation & Task: Inventory export & gap=implicit-response-validation \\
E2-LS5-T5 & E2-LS5 & adversarial / evaluation & Task: User export during field migration & gap=trap-blind-200-trust; traps=generic\_trap \\
E2-LS5-T6 & E2-LS5 & composition / evaluation & Task: Primary/backup product merge & gap=composition-validation-fallback-merge; composition=cross\_family; requires E2-LS1.pre-call-parameter-validation, E2-LS5.response-validation-fallback \\
\bottomrule
\end{longtable}
\endgroup

\subsection{E3: Data Processing \& Structured Query}
\begingroup
\scriptsize
\setlength{\tabcolsep}{1.5pt}
\renewcommand{\arraystretch}{1.05}
\settasktabwidth
\begin{longtable}{@{}%
>{\raggedright\arraybackslash}p{0.08\tasktabwidth}%
>{\raggedright\arraybackslash}p{0.08\tasktabwidth}%
>{\raggedright\arraybackslash}p{0.12\tasktabwidth}%
>{\raggedright\arraybackslash}p{0.33\tasktabwidth}%
>{\raggedright\arraybackslash}p{0.39\tasktabwidth}%
@{}}
\caption{Task design catalog for E3: Data Processing \& Structured Query.}\label{tab:task_catalog_e3}\\
\taskcatalogheader
\endfirsthead
\taskcatalogheader
\endhead
E3-LS1-T1 & E3-LS1 & canonical / learning & Task: Repair The ERP Sales Export Pipeline & gap=none \\
E3-LS1-T2 & E3-LS1 & enriched / learning & Task: Stabilize The German Revenue Export Loader & gap=gap1-invisible-characters-and-encoding \\
E3-LS1-T3 & E3-LS1 & variant / learning & Task: Repair The Revenue Normalization Job & gap=gap2-mixed-column-types \\
E3-LS1-T4 & E3-LS1 & context-shift / evaluation & Task: Repair The Customer Revenue Join & gap=implicit-schema-check-before-join \\
E3-LS1-T5 & E3-LS1 & adversarial / evaluation & Task: Fix The Sales Metric Extractor & gap=shortcut-case-sensitive-columns; traps=shortcut \\
E3-LS1-T6 & E3-LS1 & composition / evaluation & Task: Rebuild The Dirty Revenue Validation Pipeline & gap=composition-clean-query-validate; composition=cross\_family; requires E3-LS1.schema-inspection-before-query, E3-LS5.result-sanity-reconciliation \\
\midrule
E3-LS2-T1 & E3-LS2 & canonical / learning & Task: Repair The Log Timeline Sorter & gap=none \\
E3-LS2-T2 & E3-LS2 & enriched / learning & Task: Repair The German Contact Sort Order & gap=gap1-locale-aware-sorting \\
E3-LS2-T3 & E3-LS2 & variant / learning & Task: Repair The Employee Ranking Sort & gap=gap2-stable-multi-key-sort \\
E3-LS2-T4 & E3-LS2 & context-shift / evaluation & Task: Build The Q3 Sales Leaderboard & gap=implicit-ranking-sort \\
E3-LS2-T5 & E3-LS2 & adversarial / evaluation & Task: Fix The Product Catalog Ordering & gap=string-numeric-natural-sort-trap; traps=generic\_trap \\
E3-LS2-T6 & E3-LS2 & composition / evaluation & Task: Rebuild The Transaction Ordering Pipeline & gap=composition-sort-dedup-aggregate; composition=cross\_family; requires E3-LS2.type-normalization-before-sort, E3-LS3.key-alignment-before-merge, E3-LS4.null-safe-filtering-aggregation \\
\midrule
E3-LS3-T1 & E3-LS3 & canonical / learning & Task: Merge Customer Profiles With Order Summaries & gap=none \\
E3-LS3-T2 & E3-LS3 & enriched / learning & Task: Merge Population And GDP By City & gap=gap1-fuzzy-match \\
E3-LS3-T3 & E3-LS3 & variant / learning & Task: Calculate Average Course Load Per Student & gap=gap2-many-to-many-fanout \\
E3-LS3-T4 & E3-LS3 & context-shift / evaluation & Task: Compare 2023 And 2024 Product Sales & gap=implicit \\
E3-LS3-T5 & E3-LS3 & adversarial / evaluation & Task: Merge Users With Transactions For Spending Totals & gap=trap-wrong-key; traps=generic\_trap \\
E3-LS3-T6 & E3-LS3 & composition / evaluation & Task: Build A Clean Customer Revenue Dataset From Three Sources & gap=composition; composition=cross\_family; requires E3-LS2.type-normalization-before-sort, E3-LS3.key-alignment-before-merge \\
\midrule
E3-LS4-T1 & E3-LS4 & canonical / learning & Task: Average Order Amount & gap=none \\
E3-LS4-T2 & E3-LS4 & enriched / learning & Task: Regional Sales Totals & gap=gap1-groupby-dropna \\
E3-LS4-T3 & E3-LS4 & variant / learning & Task: Employee Statistics & gap=gap2-sentinel-values \\
E3-LS4-T4 & E3-LS4 & context-shift / evaluation & Task: Channel Conversion Rates & gap=implicit-null-denominator \\
E3-LS4-T5 & E3-LS4 & adversarial / evaluation & Task: Temperature Analysis & gap=trap-fillna-zero; traps=generic\_trap \\
E3-LS4-T6 & E3-LS4 & composition / evaluation & Task: Unified Product Inventory & gap=composition-multi-source-null; composition=cross\_family; requires E3-LS3.key-alignment-before-merge, E3-LS4.null-safe-filtering-aggregation \\
\midrule
E3-LS5-T1 & E3-LS5 & canonical / learning & Task: Electronics Catalog Extract & gap=baseline-rowcount-sanity \\
E3-LS5-T2 & E3-LS5 & enriched / learning & Task: Quarterly Revenue Sanity Review & gap=statistical-anomaly-detection \\
E3-LS5-T3 & E3-LS5 & variant / learning & Task: Revenue Cross-Check by Product Line & gap=cross-query-consistency \\
E3-LS5-T4 & E3-LS5 & context-shift / evaluation & Task: March Sales Report & gap=implicit-completeness-check \\
E3-LS5-T5 & E3-LS5 & adversarial / evaluation & Task: Active Customer Count & gap=logic-correctness-trap; traps=generic\_trap \\
E3-LS5-T6 & E3-LS5 & composition / evaluation & Task: Operations Quality Report & gap=composition-multi-layer-sanity; composition=cross\_family; requires E3-LS1.schema-inspection-before-query, E3-LS2.type-normalization-before-sort, E3-LS5.result-sanity-reconciliation \\
\bottomrule
\end{longtable}
\endgroup

\section{Implementation-Specific Experiment Settings}
\label{app:experiment_settings}

This appendix records implementation-specific settings needed to
reproduce the experimental protocol. It complements the main-text
description of the variants by specifying how models and provider
endpoints are routed, how execution trajectories are compacted before
being reused as evidence, and the exact Skill Author prompts used for
zero-shot generation, experience-based induction, and revision.

\subsection{Model and API Configuration}
\label{app:model_api_configuration}

We separate experimental conditions from model/API routing. The baseline and
strategy files specify the memory mechanism, retrieval behavior, and skill
authoring policy. Model-specific execution details are loaded from a model
preset file in \texttt{configs/models/*.yaml}. This keeps the same experimental
variant runnable across different providers without changing the benchmark
logic.~\Cref{tab:model_api_config_params} summarizes the configuration
parameters used for this routing.

There are two LLM call surfaces. The first is the task-solving agent, which is
launched inside Harbor through an agent adapter. The second is the host-side
LLM interface used by components such as the Skill Author and LLM-based
retriever. When a model preset is provided, both surfaces are routed through
the same preset-specific provider configuration. API secrets are not stored in
configuration files; the configs store only environment-variable names.

The provider-specific routing is explicit in the model preset. OpenAI-family
models in our runs are served through Azure OpenAI rather than the public
OpenAI endpoint. The task-solving agent uses the Codex adapter, while the
endpoint and key are forwarded through the OpenAI-compatible environment
variables expected by the adapter:

\begin{verbatim}
provider: azure_openai
harbor_agent_name: codex
agent_model_name: openai/<azure-deployment-name>

agent_env:
  OPENAI_BASE_URL: "${AZURE_OPENAI_ENDPOINT}"
  OPENAI_API_KEY: "${AZURE_OPENAI_API_KEY}"
  CODEX_MODEL_PROVIDER: azure_openai
  CODEX_PROVIDER_ENV_KEY: OPENAI_API_KEY
  CODEX_WIRE_API: responses
  CODEX_MODEL_VERBOSITY: medium

host_litellm:
  model: openai/<azure-deployment-name>
  api_base_env: AZURE_OPENAI_ENDPOINT
  api_key_env: AZURE_OPENAI_API_KEY
\end{verbatim}

Thus the \texttt{openai/...} model string denotes the OpenAI-compatible model
identifier used by Codex and LiteLLM, while the actual endpoint is the Azure
OpenAI resource named by \texttt{AZURE\_OPENAI\_ENDPOINT}. The
\texttt{agent\_model\_name} must match the Azure deployment name available in
that resource.

Claude-family models are served through Amazon Bedrock. The task-solving agent
uses the Claude Code adapter in Bedrock mode. Bedrock requires the AWS
cross-region inference-profile identifier as the task-agent model name; the
host-side LiteLLM model uses the corresponding \texttt{bedrock/...} prefix:

\begin{verbatim}
provider: bedrock
harbor_agent_name: claude-code
agent_model_name: us.anthropic.<bedrock-inference-profile-id>

agent_env:
  CLAUDE_CODE_USE_BEDROCK: "1"
  ANTHROPIC_DEFAULT_OPUS_MODEL: us.anthropic.<...>      # Opus presets
  ANTHROPIC_DEFAULT_SONNET_MODEL: us.anthropic.<...>    # Sonnet presets

host_litellm:
  model: bedrock/us.anthropic.<bedrock-inference-profile-id>
\end{verbatim}

For Bedrock runs, the host-side client relies on ambient AWS credentials, such
as \texttt{AWS\_BEARER\_TOKEN\_BEDROCK} and \texttt{AWS\_REGION}; no
\texttt{api\_base\_env} or \texttt{api\_key\_env} is stored in the model
preset. Gemini presets use \texttt{provider: gemini} with
\texttt{harbor\_agent\_name: gemini-cli}, an \texttt{agent\_model\_name} of
the form \texttt{google/<model>}, and \texttt{GEMINI\_API\_KEY} for
host-side LiteLLM calls. Kimi presets use an OpenAI-compatible Bedrock-Mantle
endpoint with \texttt{provider: bedrock\_mantle},
\texttt{harbor\_agent\_name: kimi-cli}, \texttt{agent\_model\_name} of the
form \texttt{openai/moonshotai.<model>}, and endpoint/key variables
\texttt{KIMI\_BEDROCK\_BASE\_URL} and \texttt{KIMI\_BEDROCK\_API\_KEY}.

\begin{table*}[!htbp]
\centering
\caption{\textbf{Model and API routing parameters.}
These parameters describe how model presets and run configs route task-agent
and host-side LLM calls. They are configuration fields rather than reported
evaluation metrics.}
\label{tab:model_api_config_params}
\small
\setlength{\tabcolsep}{5pt}
\renewcommand{\arraystretch}{1.12}
\begin{tabular}{@{}p{0.27\textwidth}p{0.67\textwidth}@{}}
\toprule
\textbf{Parameter} & \textbf{Meaning} \\
\midrule
\texttt{provider}
& Backend family used by a model preset for routing and documentation, e.g.,
Bedrock, Azure OpenAI, Gemini, or an OpenAI-compatible endpoint. \\

\texttt{harbor\_agent\_name}
& Agent adapter selected for task-solving trials. This determines which
preinstalled CLI or agent wrapper Harbor launches for the benchmark task. \\

\texttt{agent\_model\_name}
& Provider-qualified model identifier passed to the task-solving agent. For
deployment-based providers, this is the deployment or inference-profile name
expected by the adapter. \\

\texttt{agent\_env}
& Environment variables exported before the task-solving agent is constructed.
These variables provide provider-specific runtime settings such as endpoint
aliases, API-key variable names, wire protocol selection, or CLI trust flags. \\

\texttt{agent\_kwargs}
& Non-secret adapter arguments persisted into the Harbor agent configuration.
For Codex-style adapters this may include fields such as \texttt{base\_url},
\texttt{provider}, \texttt{env\_key}, \texttt{wire\_api},
\texttt{api\_version}, and \texttt{verbosity}. \\

\texttt{model\_name}
& The task-agent model field in the resolved baseline config. When a model
preset is used, this value is overwritten by \texttt{agent\_model\_name}. \\

\texttt{api\_base}
& Optional endpoint override in the run-level LLM config. If unset, the
provider default or preset-specific endpoint is used. \\

\texttt{api\_key\_env\_var}
& Name of the environment variable from which an API key is read. The key
value itself is never serialized into benchmark configs. \\

\texttt{host\_litellm.model}
& Model identifier used for host-side LiteLLM calls, including Skill Author
calls and LLM-based retrieval when enabled. \\

\texttt{host\_litellm.api\_base\_env}
& Environment variable whose value supplies the host-side LiteLLM API base.
This is used for provider endpoints such as Azure OpenAI or other
OpenAI-compatible gateways. \\

\texttt{host\_litellm.api\_key\_env}
& Environment variable whose value supplies the host-side LiteLLM API key.
For providers that use ambient credentials, this field can be omitted. \\

\texttt{SEVB\_HOST\_LITELLM\_MODEL}
& Runtime sentinel set by the launcher from \texttt{host\_litellm.model}. It
ensures host-side LLM components use the same preset-selected model routing. \\

\texttt{SEVB\_HOST\_LITELLM\_API\_BASE}
& Runtime sentinel containing the resolved host-side API endpoint, when an
explicit endpoint is required. \\

\texttt{SEVB\_HOST\_LITELLM\_API\_KEY}
& Runtime sentinel containing the resolved host-side API key. This value is
read from the configured environment variable at launch time. \\

\texttt{author\_model}
& Fallback model for Skill Author calls when no model preset is active. Under
preset-based runs, host-side calls are routed through the preset model instead. \\

\texttt{author\_temperature}
& Sampling temperature for Skill Author generation. \\

\texttt{author\_max\_tokens}
& Maximum generation length for Skill Author calls. \\

\texttt{judge\_model}
& Backward-compatible strategy field for selector/judge-style workflows. In
the chain strategy used by the main runs, the Skill Author path does not rely
on a separate judge model. \\

\texttt{judge\_temperature}
& Sampling temperature for judge/selector calls when that strategy path is
enabled. \\

\texttt{judge\_max\_tokens}
& Maximum generation length for judge/selector calls when that strategy path
is enabled. \\

\texttt{default\_embed\_model}
& Default embedding model used by embedding-based retrieval if no specialized
embedder is provided. \\

\texttt{harbor\_orchestrator\_type}
& Harbor execution backend. The config supports local and remote
orchestrators, but this field changes infrastructure rather than the
benchmark protocol. \\

\texttt{harbor\_n\_concurrent\_trials}
& Number of concurrent Harbor trials. The lifelong protocol requires this to
be \texttt{1}, because task order and library state must remain sequential. \\
\bottomrule
\end{tabular}
\end{table*}

At launch time, the model preset first resolves \texttt{agent\_env} against
the process environment. It then sets \texttt{harbor\_agent\_name} and
\texttt{model\_name} in the baseline config used by Harbor. For host-side
calls, the launcher publishes the resolved \texttt{host\_litellm} settings
through \texttt{SEVB\_HOST\_LITELLM\_*} variables. The runtime reads these
sentinels before falling back to the run-level \texttt{api\_base},
\texttt{api\_key\_env\_var}, and strategy-level \texttt{author\_model}
defaults.

\subsection{Trajectory Compaction}
\label{app:trajectory_compactor}

After each Harbor trial, we persist the verifier result and a compacted view of
the agent trajectory. The compactor is deterministic and does not call an LLM.
It parses the recorded trajectory, selects a small set of informative events,
and renders them as a bounded markdown summary. We use two compaction modes:
\emph{rich} compaction for Skill Author evidence, and \emph{rough}
compaction for the raw-trajectory control baseline. \Cref{tab:trajectory_compactor_settings}
lists the parameters for both modes.

The rich compaction is stored as \texttt{trajectory\_compact}. It preserves
agent messages, tool calls, observations, verifier summaries, and available
\texttt{reasoning\_content} when the harness records it. This field is used by
experience-based induction and revision. For induction, the Skill Author sees
the current canonical-task compacted trajectory. For revision, the strategy
constructs a cumulative same-family learning history by concatenating prior
learning-role compacted trajectories with the current trial. Deployment and
replay trials are not used as authoring evidence.

The rough compaction is stored as \texttt{trajectory\_compact\_rough}. It is
used only by \textsc{Raw-Trajectory}. Rough compaction intentionally drops
\texttt{reasoning\_content}, because the raw-trajectory baseline should expose
episodic execution evidence rather than the agent's private abstraction. The
baseline retrieves same-family learning-role trajectories only, in chronological
order, with \texttt{trajectory\_retrieval\_k=3}. Thus canonical tasks have no
past trajectories, enriched tasks can see the canonical trace, variant tasks
can see canonical and enriched traces, and deployment tasks can see the full
canonical--enriched--variant chain.

\begin{table*}[t]
\centering
\caption{\textbf{Trajectory compactor settings.}
Both modes are deterministic. Rich compaction is the evidence source for the
Skill Author; rough compaction is the evidence source for the raw-trajectory
baseline.}
\label{tab:trajectory_compactor_settings}
\small
\setlength{\tabcolsep}{5pt}
\renewcommand{\arraystretch}{1.12}
\begin{tabular}{@{}p{0.27\textwidth}p{0.32\textwidth}p{0.32\textwidth}@{}}
\toprule
\textbf{Parameter} & \textbf{Rich Compaction} & \textbf{Rough Compaction} \\
\midrule
\texttt{head}
& Keeps the first \texttt{8} events.
& Keeps the first \texttt{4} events. \\

\texttt{tail}
& Keeps the last \texttt{12} events.
& Keeps the last \texttt{8} events. \\

\texttt{max\_event\_chars}
& \texttt{8000}.
& \texttt{8000}. \\

\texttt{max\_total\_tokens}
& \texttt{8000} approximate tokens per trial.
& \texttt{8000} approximate tokens per trial. \\

\texttt{max\_reasoning\_chars}
& \texttt{None}; preserve recorded reasoning content subject to the
per-event cap.
& \texttt{0}; drop reasoning content entirely. \\

\texttt{max\_message\_chars}
& \texttt{None}; preserve agent messages subject to the per-event cap.
& \texttt{None}; preserve agent messages subject to the per-event cap. \\

\texttt{max\_observation\_chars}
& \texttt{3000}.
& \texttt{3000}. \\

\texttt{max\_tool\_args\_chars}
& \texttt{200}.
& \texttt{100}. \\

\texttt{signal} events
& Always keeps events matching traceback, error/exception/failure markers,
\texttt{/skills/}, \texttt{reward.txt}, or verifier output.
& Same signal-event rule. \\
\bottomrule
\end{tabular}
\end{table*}

The compactor supports several trajectory formats. It first parses the
canonical ATIF-style \texttt{trajectory.json} with \texttt{steps}; if that is
not available, it falls back to common JSON shapes such as \texttt{events},
\texttt{messages}, \texttt{trace}, top-level lists, JSONL logs, and finally
plain text. For each selected event, the rendered summary records the source
kind, message text, optional reasoning content, up to three tool calls,
observations or errors, and the verifier verdict with reward. It also extracts
referenced skill paths so later analysis can audit whether the agent actually
used mounted skills.

%

\subsection{Skill Author Prompts}
\label{app:skill_prompts}

The Skill Author is a host-side LLM call used only for skill creation and
revision. The task-solving agent never receives these prompts. The
implementation separates each call into a stable system prompt and a
trial-specific user prompt. The system prompt contains the reusable skill
writing contract, while the user prompt supplies the family id, skill id,
verifier feedback, trajectory summary, existing skills, and condition-specific
instruction. The exact implementation prompt code is reproduced below.

\lstdefinestyle{sevbpromptstyle}{
  basicstyle=\ttfamily\scriptsize,
  breaklines=true,
  breakatwhitespace=false,
  columns=fullflexible,
  keepspaces=true,
  showstringspaces=false,
  tabsize=2,
  language=Python,
  frame=single,
  framerule=0.35pt,
  rulecolor=\color{black!45},
  backgroundcolor=\color{gray!2},
  literate=
    {←}{{$\leftarrow$}}1
    {→}{{$\rightarrow$}}1
    {↔}{{$\leftrightarrow$}}1
    {≠}{{$\ne$}}1
    {≥}{{$\ge$}}1
    {─}{{-}}1
    {├}{{|}}1
    {└}{{|}}1
}

\paragraph{Prompt Rendering Logic.}
The Skill Author uses the same stable system contract across authoring calls,
paired with a task-specific user message. Revision prompts receive the
revision mode, verifier outcome, pass/fail-specific diagnosis instruction,
cumulative same-family learning trajectories, and the current same-family
skills that may be edited. Experience-based induction prompts receive the
family id, latent skill id, canonical-task verifier feedback, and the compacted
canonical trajectory. Zero-shot prompts receive only family metadata and the
target skill id, and therefore explicitly prohibit invented task-specific
gotchas or Tier-3 resources.

\paragraph{Prompt Templates.}
The following source excerpt contains the shared system contract and the exact
system/user templates for revision, experience-based induction, and zero-shot
generation. The listings are split into smaller blocks to avoid LaTeX memory
pressure from one very large code box.

\noindent\textbf{Shared skill-writing contract}\par\smallskip
\begin{lstlisting}[style=sevbpromptstyle]
# three questions the LLM should keep in mind while authoring:
#   1. What is a skill (vs a task note)?
#   2. What are the three tiers (Tier 1 / Tier 2 / Tier 3)?
#   3. What mindset produces a good skill -- comprehensive, adversarial-aware,
#      written for the FUTURE agent rather than the current trace.
# Kept short on purpose; the detailed limits + best practices live in
# _SPEC_BLOCK below. Both are concatenated into every system prompt.
_SKILL_MINDSET = """\
================================================================================
WHAT A SKILL IS, AND WHAT YOU ARE BUILDING
================================================================================

A skill is a *reusable capability package* persisted on disk and loaded into
a future agent's context whenever it tackles a task in this domain. It is
NOT a one-off note about today's trial -- it must generalize to the whole
class of similar tasks, including ones the current trace did not cover.

## The three tiers (agentskills.io progressive disclosure)

  Tier 1 -- METADATA: the YAML frontmatter at the top of SKILL.md
            (`name` + `description`). ~100 tokens, ALWAYS in the agent's
            context. The description is the entire trigger surface --
            future tasks RETRIEVE this skill solely on its match against
            user prompts. Weak description = the skill never fires.

  Tier 2 -- INSTRUCTIONS: the SKILL.md body (markdown, after the
            frontmatter). Loaded ONLY when Tier 1 matches. <5000 tokens
            recommended. This is the workflow, gotchas, decision rules,
            anti-patterns, and worked examples a future agent will follow.

  Tier 3 -- RESOURCES: optional files in `<slug>/scripts/`,
            `<slug>/references/`, `<slug>/assets/`. Loaded ON DEMAND when
            SKILL.md explicitly tells the agent to run / read / copy them.
            Tier 3 holds executable validators (`scripts/`), bulky
            reference docs (`references/`), and copy-paste templates
            (`assets/`) -- material that would bloat Tier 2 if inlined.
            Uncited Tier-3 files are dead weight and rejected.

## Mindset: what makes a good skill

* **Aim for COMPREHENSIVE coverage.** The trial in front of you is one
  observation. Design the skill so it also handles the obvious harder
  variants a future task will throw -- different data shapes, missing or
  malformed fields, partial failures, cross-cutting edge cases, scaled-up
  inputs. Encode the GENERAL workflow, not just what worked today.

* **Anticipate ADVERSARIAL conditions.** Future agents will see tasks
  designed to expose weaknesses: misleading framing, edge cases that
  look like the canonical case, traps that punish "blindly follow the
  workflow" patterns, inputs crafted to make a tempting-but-wrong path
  look correct. Encode the GUARDRAILS -- validation steps, sanity
  checks, "do not rely on X when Y" anti-patterns -- that catch these
  before the agent commits to a wrong answer.

* **Write for the AGENT in the FUTURE, not for the human reading this
  trace.** The future agent will encounter cases neither you nor the
  trace has seen. Ask: what knowledge would let it succeed where it
  would otherwise fail or guess? Cache that knowledge -- runtime paths,
  schema names, exact APIs, conventions, gotchas -- so the future
  agent does not have to RE-DERIVE it.

* **Bias toward DECISION RULES** ("if X pattern appears, do Y instead
  of Z"; "before doing W, validate V"; "never call A when B is true").
  Decision rules generalize. Generic prose advice ("be careful",
  "consider edge cases") is filler.

* **Keep it CONCISE.** Comprehensive ≠ verbose. Every token in Tier 2
  is paid context for every future task. Add only what is non-obvious,
  reusable, and worth its cost. Move bulky detail to Tier 3.

* **PROCEDURAL knowledge, not EPISODIC memory.** A skill encodes
  HOW to handle a CLASS of problems. It does NOT record the specific
  answers, fixes, or trajectory steps from any single trial. WRITE:
  transferable failure patterns, reusable workflows, trigger
  conditions, validation checks, decision rules. DO NOT WRITE: the
  hard-coded fix for one task instance, a literal answer to one
  problem, the full log of a single trajectory, "in the FinGuard
  task we did X". The skill must work for tasks neither you nor the
  trace has seen.

The detailed agentskills.io spec follows; treat it as the binding
contract for shape (directory layout, frontmatter fields, Tier 3
folders, etc.).

"""


# Distilled from the agentskills.io spec, best-practices, optimizing-descriptions,
# and using-scripts pages. Embedded into every prompt so the LLM author has the
# full design contract without an external lookup.
_SPEC_BLOCK = """\
================================================================================
AGENT SKILLS -- SPEC + BEST PRACTICES (READ BEFORE WRITING)
================================================================================

## 1. Directory layout

A skill is a directory; minimum content is `SKILL.md`. Optional subdirs are
`scripts/`, `references/`, and `assets/`:

```
my-skill/
├── SKILL.md          # Required: metadata + instructions
├── scripts/          # Optional: executable code
├── references/       # Optional: documentation
├── assets/           # Optional: templates, resources
└── ...               # Any additional files or directories
```

Reference these from SKILL.md by RELATIVE path (e.g. `scripts/extract.py`,
`references/REFERENCE.md`). Keep file references one level deep.

## 2. SKILL.md format = YAML frontmatter + Markdown body

Frontmatter is the YAML between `---` markers at the very top.

| Field           | Req | Constraint                                                                           |
|-----------------|-----|--------------------------------------------------------------------------------------|
| `name`          | YES | <=64 chars; [a-z0-9-]; no leading/trailing/consecutive `-`; MUST match folder name. |
| `description`   | YES | <=1024 chars; non-empty; states WHAT the skill does AND WHEN to use it.              |
| `license`       | no  | License name or path to a bundled license file.                                      |
| `compatibility` | no  | <=500 chars; environment requirements (product, packages, network, runtime version). |
| `metadata`      | no  | Free key-value map; pick keys unlikely to collide.                                   |
| `allowed-tools` | no  | Space-separated pre-approved tools (experimental).                                   |

Minimal valid SKILL.md:

```
---
name: pdf-processing
description: Extract PDF text, fill forms, merge files. Use when handling PDFs.
---
# PDF processing
... body ...
```

The Markdown body has no enforced format. Recommended sections: step-by-step
instructions, examples, common edge cases. The full body loads into the agent
once the skill activates -- treat every token as paid context.

## 3. Progressive disclosure (THREE TIERS)

Skills load in three tiers; design for it.

  Tier 1 -- METADATA (~100 tokens, ALWAYS in context): the YAML frontmatter
            (`name` + `description`). Used by the agent's matcher to decide
            whether to load the body. A weak description = the skill never
            triggers.
  Tier 2 -- INSTRUCTIONS (<5000 tokens recommended, loaded WHEN ACTIVATED):
            the SKILL.md body. Keep it under 500 lines. Move bulky detail to
            Tier 3.
  Tier 3 -- RESOURCES (loaded ON DEMAND): files in `scripts/`, `references/`,
            `assets/`. The agent reads / runs these only when SKILL.md tells
            it to. Use this tier for big artefacts, executables, rarely
            needed detail. Cite each Tier-3 file from the body with a clear
            "when to load / run it" trigger; uncited Tier-3 files are dead
            weight.

## 4. Writing the description (Tier 1, highest leverage)

The description's CORE PURPOSE is to TRIGGER skill loading -- not to
introduce or sell the skill. The agent's matcher reads it against the
incoming user prompt; if it doesn't match, none of your Tier-2 / Tier-3
work matters because the skill never fires.

### Required ingredients (every description should contain ALL):

1. **Task object(s)** -- the concrete things the skill operates on.
   Examples: `Dockerfile`, `Express middleware`, `JSON schema`,
   `LaTeX table`, `pytest fixture`, `verifier output`, `npm package`,
   `bearer token`, `trajectory.json`. List 2-4.
2. **Action verb(s)** -- what the skill DOES with those objects.
   Examples: `inspect`, `debug`, `validate`, `summarize`, `patch`,
   `compare`, `extract`, `aggregate`, `refactor`, `migrate`. List 2-4.
3. **WHEN scenarios** -- 2-3 concrete trigger contexts in
   "Use this skill when..." form. Include INDIRECT phrasings users
   actually type ("the test is timing out", "the verifier reports
   X", "the JSON file in /tmp doesn't validate"). Synonyms count.
4. **Optional: boundary conditions** -- "Do not use this skill for
   X / Y" lines. Reduces false-positive triggers when neighbouring
   skills exist.

### Style:

- **Third-person / imperative voice**: "This skill should be used
  when..." or "Use when...". NEVER "I help with..." / "you can use
  this to...".
- **<=1024 chars hard cap.** Aim for 4-8 lines.
- **Cover INDIRECT triggers** (the user does not name the domain):
  "the failing test mentions <error>", "the spreadsheet attached",
  "the file in /tmp".

### Banned patterns (REJECTED):

- Generic catch-all words as the only signal: `debug`, `fix`,
  `code`, `bug`, `task`, `improve`, `helper`, `assist`. These match
  everything = they match nothing.
- Meta-talk about the skill itself: `"a structured workflow for..."`,
  `"this skill helps with..."`, `"records past experience"`,
  `"saves the agent from..."`. Dilutes keyword signal.
- Self-referential phrasing: `"This skill records what we learned
  in trial X"`. Skills describe TASK DISTRIBUTIONS and TRANSFERABLE
  OPERATIONS, not their own provenance.
- Walking the body: don't put the workflow steps in the description.
  Description = WHEN to fire; body = WHAT to do.

### Concrete examples:

  GOOD (auth-bypass family in Express/Node):
    "Diagnose and fix bearer-token / JWT authentication-bypass bugs
    in Node.js / Express middleware where invalid or malformed
    tokens leak protected user data. Use when an auth middleware
    (`tokenValidator`, `roleChecker`, `tokenPolicy`) defers errors
    instead of returning 401, when `/api/users`-style routes return
    user records under bad tokens, when soft-fail / monitor-only
    auth modes mask 401-worthy failures, or when the verifier
    reports user-data fields (`email`, `phone`, `ssn_last4`)
    leaking through 200 responses. Do not use this skill for plain
    role/RBAC bugs where authentication itself is correct."

  POOR (does NOT trigger reliably):
    "Use when debugging a software bug. Provides a structured,
    step-by-step workflow for identifying the root cause."
    -- generic; no task object, no domain keyword; matches every
    coding task = matches none in practice.

## 5. Body best practices

The body is what the agent reads AFTER the description triggers. Its
job is PROCEDURAL: tell the agent what to do, in order, with
explicit branches. It is NOT a paper, a textbook chapter, or a
narrative summary.

### What the body MUST cover (when applicable):

- **Goal statement** -- one sentence: what does success look like.
- **When to use** -- echo the description's WHEN scenarios, more
  concretely. May include "do not use when..." boundaries.
- **Workflow** -- numbered or `- [ ]`-checklist of action steps in
  order. Each step starts with an action verb ("First inspect...",
  "Then verify...", "If X, do Y; else do Z."). NOT prose paragraphs.
- **Gotchas** -- concrete near-miss / failure patterns the agent
  will hit otherwise (soft-deletes, ID renames, health checks that
  lie, off-by-one timezone drift). Decision-rule form.
- **Resource map** -- a short table or list mapping triggers to
  Tier-3 files: "Run `scripts/X.py` when ...", "Read
  `references/Y.md` if ...", "Use `assets/Z.tmpl` to format ...".
  REQUIRED whenever the skill has Tier-3 files.
- **Output format / completion criteria** -- when the task expects
  a specific output shape, embed a literal template (agents
  pattern-match against concrete structures better than prose).
- **Uncertainty handling** -- "if information X is missing, do Y
  before proceeding" / "if validator output is ambiguous, default
  to the SAFER branch". Tells the agent how to act under partial
  information rather than guessing.

### Style requirements:

- **Action-oriented language.** Every step starts with a verb in
  imperative or "First / Then / If / Avoid" form. NO sentences like
  "It is important to note that...".
- **Decision rules over advice.** "If <pattern>, do <action>" beats
  "consider edge cases" / "be careful".
- **Defaults over menus.** Pick ONE recommended path; relegate
  alternatives to a brief "Alternatives" line. Forcing the agent to
  re-decide every call wastes tokens.
- **Match specificity to fragility.** PRESCRIPTIVE on fragile /
  destructive / order-sensitive steps; PERMISSIVE on creative
  steps. Mixing both within one SKILL.md is fine.
- **One worked example** beats exhaustive documentation. Trust the
  agent's general competence on syntax / language basics.
- **Validation loop.** When the workflow is iterative: do-work,
  run-validator (a `scripts/` file or shell command), fix, repeat.
  State the validator command explicitly.
- **Plan-validate-execute** for batch / destructive ops: agent
  produces a structured plan, validates against a source of truth,
  THEN executes.

### What the body MUST NOT contain:

- Background essays / theory dumps that the agent already knows
  (basic Python syntax, "what is JWT", "what is Docker").
- A literal log of any single trial's trajectory.
- Hard-coded answers / fixes for one task instance ("change line
  43 of tokenValidator.js to..." for a specific repo). Encode the
  PATTERN ("when soft-fail logic defers auth errors, remove the
  defer branch and return 401 + log") instead.
- Long enumerations (full enum tables, exhaustive API listings) --
  move those to `references/<topic>.md`.
- Resources without a citation. If you bundle `scripts/foo.py` you
  MUST tell the agent in the body: WHEN to run it, exact command,
  expected output, how to interpret failure.

### The body must be READABLE COLD by a future agent.

Assume the agent has zero memory of any past trial. Don't write
"as before" / "as in the previous task" / "the same fix as last
time". Every reference must be self-contained.

## 6. Tier-3 directories

The agent's mental model: it RUNS files in `scripts/`, READS files in
`references/`, COPIES files in `assets/`. Place each file by its
intended interaction.

Every Tier-3 file MUST be cited from SKILL.md with a clear
"when to run / read / copy" trigger. Uncited Tier-3 files are dead
weight and rejected.

Naming (all three dirs):
  - Specific verb_object pattern preferred:
    `validate_skill_frontmatter.py`, `extract_failure_modes.py`,
    `compare_skill_versions.py`, `failure_patterns.md`,
    `result_schema.json`, `report_template.md`.
  - REJECTED: vague names like `helper.py`, `notes.md`, `more.md`,
    `stuff.txt`, `template2.md`, `old.json`, `doc1.md`.

---

`scripts/` -- EXECUTABLE CODE the agent RUNS.

What goes here: deterministic, repeatable operations. Examples for
benchmarking-style domains: `parse_trajectory.py`,
`extract_failure_modes.py`, `validate_skill_frontmatter.py`,
`check_skill_patch.py`, `summarize_verifier_logs.py`,
`aggregate_results.py`, `compare_quotes.py`.

When to bundle one: trace shows the agent RE-DERIVING the same
logic, OR the step is fragile and a tested script beats freeform
generation.

Hard rules (REJECTED otherwise):
  - **CLI**: clear arg interface; `--help` with brief description +
    flags + examples.
  - **Output**: structured (JSON / CSV / TSV) on stdout;
    diagnostics on stderr; deterministic given same inputs.
  - **Errors**: explain WHAT went wrong, WHAT was expected, WHAT to
    try. NOT bare tracebacks.
  - **Exception handling** for common failure modes: missing file,
    malformed JSON / YAML, schema invalid, missing required field.
  - **Idempotent** (agents retry).
  - **Closed-set / enum** params for ambiguous choices.
  - **`--dry-run`** for destructive ops; non-default for any
    deletion / overwrite. NEVER default to deleting files,
    overwriting results, or clearing dirs.
  - **No interactive prompts**; agents run non-interactive shells.
  - **No silent network installs**; declare deps inline (PEP 723
    `# /// script` for Python; `npm:` for Deno; etc.) or in
    SKILL.md `compatibility` field.
  - **Meaningful exit codes**, documented in `--help`.
  - **Predictable output size**; support `--output FILE` for large.

SKILL.md must specify, per script: WHEN to run, exact command,
expected output shape, how to interpret failure.

---

`references/` -- ON-DEMAND DOCUMENTATION the agent READS.

What goes here: long-form material that would bloat SKILL.md if
inlined. Each file serves ONE clear topic. Examples for a
benchmarking / self-evolving domain:
  - `failure_patterns.md` -- catalogue of TRANSFERABLE failure modes
    (NOT single-trial details).
  - `update_policy.md` -- when a skill update is allowed vs.
    rejected.
  - `procedural_skill_rubric.md` -- what makes a skill "good".
  - `benchmark_task_schema.md` -- task spec / verifier interface.
  - `trajectory_analysis_protocol.md` -- step-by-step trace review.
  - `skill_patch_rubric.md` -- patch acceptance criteria.

Conventions:
  - One topic per file. Filename describes the topic in 2-4 words.
  - Long files (>200 lines): start with a TOC.
  - SKILL.md links each reference DIRECTLY: "Read
    `references/failure_patterns.md` if the verifier reports
    pattern Y". NO daisy-chained jumps (ref → ref → ref).
  - NO `references/all.md` dumping ground.
  - NO obviously expired or task-instance-specific content unless
    explicitly marked.

---

`assets/` -- STATIC RESOURCES the agent COPIES / FILLS / EMBEDS
verbatim (never executed, rarely "read for understanding").

What goes here: document templates, config skeletons, prompt
scaffolds, lookup tables, JSON / YAML schemas, LaTeX templates,
sample input/output pairs. Examples:
  - `skill_template.md`
  - `skill_patch_template.md`
  - `result_schema.json`
  - `evaluation_table_template.tex`
  - `report_template.md`

Conventions:
  - Files are STRUCTURED, REUSABLE, READY to drop in with minimal
    edits.
  - SKILL.md tells the agent WHEN to use each: "Use
    `assets/report_template.md` as the output structure; replace
    `{{placeholders}}`".
  - REJECTED here: long manuals (those go to `references/`),
    backup / dead files, random notes, unused data.

## 7. One-off commands (no `scripts/` needed)

If an existing tool already does the job, reference it directly with a
pinned version: `uvx ruff@0.8.0 check .`, `npx eslint@9 --fix .`, etc.
State runtime prerequisites in SKILL.md or in the `compatibility`
frontmatter field.

## 8. Self-evolving skill discipline (HARD)

You are part of an evolving library: skills get revised after each
trial. That makes the line between "transferable knowledge" and
"trial-specific noise" critical. Discipline:

### What a skill MUST be:

- A description of a TASK DISTRIBUTION + the TRANSFERABLE OPERATIONS
  that solve it.
- A workflow that works on tasks the current trace did NOT cover.
- A library of failure patterns + decision rules that generalize.

### What a skill MUST NOT be:

- A log of any single trajectory.
- A specific answer to one task instance ("the fix is to delete
  line 43 of `tokenValidator.js`").
- A scrapbook of "what we learned in trial X".

### Update policy (applies to every revise / propose call):

- **Update when** the trace shows a failure pattern that is likely
  REPEATABLE across the family, AND the fix can be encoded as a
  decision rule / workflow step / Tier-3 artefact that future agents
  can apply WITHOUT having seen this trial.
- **Do NOT update** for a one-off quirk that only this task instance
  exhibits. Silently accept the existing skill (NoOp-equivalent
  revise: return SKILL.md verbatim).
- **Do NOT inflate** the body just to "have something to say".
  If the trace teaches nothing new, NoOp.
- **NEVER write task-specific answers** into SKILL.md / references/
  / assets/. That pollutes the library and biases future tasks.

### Patch self-review checklist (run mentally before emitting JSON):

- [ ] Is every change driven by a pattern that will recur, not by
      this trial's specific symptoms?
- [ ] Could a future agent, given only the new SKILL.md (no trace),
      apply the workflow to a DIFFERENT task in this family?
- [ ] Is the description still a strong trigger (≥5 domain
      keywords; concrete WHEN scenarios)?
- [ ] Does every Tier-3 file have a clear "when to run / read /
      copy" trigger in SKILL.md?
- [ ] Did I avoid copy-pasting trajectory text or task-instance
      identifiers (file paths specific to this repo, exact line
      numbers, the literal failing input)?

If any answer is "no", revise the patch before submitting.
================================================================================
"""
\end{lstlisting}

\noindent\textbf{Revision prompt templates}\par\smallskip
\begin{lstlisting}[style=sevbpromptstyle]
_REVISE_SYSTEM = """\
You are a skill-evolution assistant maintaining a shared library of
procedural skills used by an LLM agent. Your role on this call is to
REVISE existing SKILL.md content (and optionally bundled Tier-3 files)
based on a recent task trial. Output STRICT JSON.

""" + _SKILL_MINDSET + _SPEC_BLOCK + """

================================================================================
HOW TO PROCESS A REVISION REQUEST
================================================================================

The user message provides, in order:
  ## Revision mode + mode-specific instruction
  ## Trial outcome (PASS or FAIL banner + verifier breakdown + diagnosis_rule)
  ## Compacted trajectory (the agent's recent run)
  ## Target skills (slug list + task family + current SKILL.md content)

Apply these workflow rules (in addition to the spec block above):

1. **Apply the diagnosis_rule** provided in the trial-outcome section.
   The PASS branch and the FAIL branch ask for different things; do NOT
   substitute one for the other. If passed and the trajectory shows
   nothing the skill can absorb, return upsert_files containing only the
   existing SKILL.md text verbatim -- the parser treats that as NoOp.

2. **Respect the layering** (see Progressive Disclosure in the spec):
   - Don't bloat SKILL.md with material that belongs in `references/`
     (long enums, edge-case catalogues, vendored API docs).
   - Don't shrink the `description` for token savings -- discoverability
     beats a few chars.
   - Don't change `name` unless operation_type=replace; renaming breaks
     the folder-slug match.
   - **`description` quality is the trigger surface** (Tier-1, always
     in future agents' context). When you revise, AUDIT the existing
     `description`. If it is generic ("Use when debugging a software
     bug.") or lacks domain keywords, REWRITE it to satisfy ALL of:
       a. <=1024 chars, imperative voice.
       b. AT LEAST 5 concrete keywords from the trajectory's domain
          (file types, tool names, error symptoms, library names,
          API verbs, domain terms). Generic words ("debug", "fix",
          "code", "bug", "task") DO NOT count.
       c. WHAT + WHEN: 2-3 concrete trigger contexts including
          INDIRECT ones (where the user does NOT name the skill
          domain).
       d. No meta-talk ("a structured workflow for...", "this skill
          helps with..."). It dilutes keyword signal.
     A weak description = the skill never fires on future tasks =
     all your other revisions are wasted.

3. **operation_type semantics**:
   - `revise`  -- normal edit; same name, same intent.
   - `narrow`  -- tighten `description` / `## When to use` so the skill no
                  longer triggers on the failing context.
   - `replace` -- full rewrite; only when minor edits cannot fix the
                  pattern.
   - `create`  -- introduces a new skill (see rule 5; must be in the
                  same family as the current task).

4. **Tier-3 files** (scripts/, references/, assets/) on EXISTING skills
   MUST be cited from SKILL.md with a clear "when to read / run / copy"
   trigger. Uncited Tier-3 files are dead weight -- the parser rejects
   them.

5. **Update existing vs create new (SAME FAMILY ONLY).**

   When the family already has skills (shown in the "Skills in family"
   block of the user message), your DEFAULT action is to SELECT ONE OR
   MORE of them and REFINE them in place. Creating a brand-new sibling
   is the exception, not the default.

   ADDITIVE BIAS for revisions:
   - PRESERVE the existing SKILL.md structure, naming, and intent
     unless the trajectory shows specific content is wrong, misleading,
     or unsafe.
   - PREFER ADDING over REPLACING: a new Gotcha, a tightened Workflow
     step, a refined `## When to use` clarifier, an extra cited Tier-3
     file, a sharper one-line summary in the frontmatter.
   - You MAY rewrite a section ONLY if the trajectory shows that
     section was wrong; cite the failing step in `summary`.
   - DO NOT shrink, gut, or wholesale-rewrite an existing SKILL.md
     "for clarity" / token savings -- accumulated knowledge IS the
     point of the library.
   - If the SKILL.md is fundamentally misaligned with the trajectory
     (rare), use operation_type=`replace` -- NEVER `create` with the
     same slug (see slug↔op pairing below).

   SLUG ↔ operation_type pairing (HARD CONSTRAINT):
   - Reusing a slug already shown in "Skills in family" → op MUST be
     `revise` / `narrow` / `replace`. NEVER `create`.
   - `create` → MUST use a brand-new slug not visible in that block.
   - Violations are REJECTED by the parser.

   You MAY create a new skill IFF the trajectory exposes a genuinely
   DISTINCT capability the existing skill is not the right place for
   (different workflow / problem space / prerequisites).

   To create a new skill in the task's family:
   a. Pick a NEW slug. Naming rules (mandatory):
      - Lowercase ASCII letters, digits, and `-` only (kebab-case)
      - Describes the capability concisely (3-6 hyphen-separated words)
      - Globally unique across ALL families. Check the existing slug
        list in the user message and avoid any visible name.
      - Do NOT include the family_id in the slug; the parser composes
        the formal id as `<task_family_id>.<your_slug>`.
   b. Add EXACTLY ONE entry to upsert_files:
        "<your_slug>/SKILL.md": "<full SKILL.md content>"
      The folder name and SKILL.md frontmatter `name` MUST match.
   c. NO Tier-3 files for new skills. The parser REJECTS Tier-3 paths
      under a new slug. Tier-3 is reserved for revisions to existing
      skills (rule 4).
   d. The new skill is recorded as `<task_family_id>.<your_slug>` --
      you cannot create a skill in another family from this task.
   e. In `summary`, state you are creating a new skill and briefly
      justify why no existing skill could absorb the evidence.

   Updating an existing skill AND creating a new skill in the same patch
   is allowed -- list both under upsert_files. Set operation_type to
   "create" if any new slug is involved, otherwise "revise" / "narrow"
   / "replace" per rule 3.

## Required JSON output

Return EXACTLY one JSON object with these keys:

```
{
  "summary": "<one short sentence describing the change>",
  "operation_type": "revise" | "narrow" | "replace" | "create",
  "upsert_files": {
    "<slug>/SKILL.md": "<full new SKILL.md content>",
    "<slug>/scripts/<file>": "<optional, only if SKILL.md cites it>",
    "<slug>/references/<file>": "<optional, only if SKILL.md cites it>",
    "<slug>/assets/<file>": "<optional, only if SKILL.md cites it>"
  },
  "delete_paths": []
}
```

Always include FULL file content -- never diffs. Include only files you
actually edited or created. No commentary outside the JSON.
"""


_REVISE_USER_TEMPLATE = """\
## Revision mode: {mode}

{mode_instruction}

## Trial outcome

{outcome_section}

## Cumulative trajectory across this family's learning trials

The trace below concatenates ALL prior learning-block trials in this
family (T1, T2, ...) plus the trial that just finished. Each chunk is
delimited by a `## Past trial:` or `## Current trial:` header. Reflect
across the full episodic history -- patterns that appear repeatedly are
stronger evidence than single-trial signals.

```
{trajectory_summary}
```

## Skills in family `{family_id_for_this_task}`

This family has {n_existing} active skill(s): {existing_slug_list}.

**Default action when skills exist: SELECT ONE OR MORE of them and
REFINE in place.** Apply the additive bias from system rule 5 --
preserve existing structure and content; ADD what the trajectory
teaches (gotchas, tightened steps, refined frontmatter summary, cited
Tier-3 files). Do NOT rewrite or shrink existing SKILL.md content
unless the trajectory proves a specific section is wrong.

The retriever suggested {slug_list} as the most relevant edit point
("← retriever suggested" marker), but you are NOT constrained to it
-- you may:

  - revise the suggested skill (additive refinement; default),
  - revise multiple skills in this family in one patch,
  - revise a DIFFERENT skill in this family that the trajectory shows
    is actually wrong / incomplete,
  - leave one skill untouched and edit another,
  - create a NEW sibling skill ONLY if the trajectory exposes a
    genuinely distinct capability the existing skills cannot absorb
    (see system prompt rule 5),
  - return all current SKILL.md content verbatim if the trajectory
    adds nothing new (NoOp-equivalent).

**REMINDER (system rule 5 hard constraint):** the slugs listed above
({existing_slug_list}) MUST NOT be reused under operation_type=`create`.
For those slugs use `revise` / `narrow` / `replace`. `create` is only
valid with a brand-new slug.

{target_blocks}
"""
\end{lstlisting}

\noindent\textbf{Experience-based induction prompt templates}\par\smallskip
\begin{lstlisting}[style=sevbpromptstyle]
_INDUCE_SYSTEM = """\
You are a skill-evolution assistant maintaining a shared library of
procedural skills used by an LLM agent. Your role on this call is to
INDUCE a NEW skill from an LLM agent's recent task experience -- this
is the FIRST version of this skill in the library. Output STRICT JSON.

""" + _SKILL_MINDSET + _SPEC_BLOCK + """

================================================================================
HOW TO PROCESS AN INDUCTION REQUEST
================================================================================

The user message provides:
  ## Family id, latent skill id, slug
  ## Last task verdict (verifier_passed + failure_summary)
  ## Compacted trajectory (your only ground-truth evidence)

Apply these induction rules (in addition to the spec block above):

1. **The trajectory is your evidence.** Treat it like a debrief:
   - Where did the agent stumble, improvise, retry, or miss the obvious?
     Those moments become Gotchas, Workflow steps, or bundled scripts.
   - What context (file paths, schema names, exact APIs, conventions)
     did the agent have to discover at runtime? Encode it directly so
     future runs do not.
   - Skip what the agent already knows (language syntax, file formats,
     standard libraries). Only project- and domain-specific knowledge
     that the agent had to RE-DERIVE.

2. **Frontmatter** -- `name` and `description` (HARD constraints):

   `name` MUST equal the slug given in the user message.

   `description` is the entire trigger surface (Tier-1, always in
   future agents' context). A weak description = the skill never
   fires. REQUIREMENTS:

   a. <=1024 chars, imperative voice ("Use this skill when...").
   b. **Keyword density**: list AT LEAST 5 concrete keywords
      extracted from the trajectory's task domain -- file types,
      tool names, error symptoms, library/framework names, API
      verbs, domain terms. Generic words ("debug", "fix", "code",
      "bug", "task") DO NOT count as keywords.
   c. **State WHAT (capability) AND WHEN (trigger contexts)**, not
      just one. The "when" should list 2-3 concrete trigger contexts
      -- including INDIRECT ones where the user does NOT name the
      skill domain ("the spreadsheet attached", "the file in /tmp",
      "the failing test mentions <error>"). Synonyms count.
   d. NO meta-talk about the skill itself ("a structured workflow
      for...", "this skill helps with...") -- agents skim
      descriptions and meta-talk dilutes keyword signal.

   GOOD example (for a hypothetical auth-bypass family):
     "Diagnose and fix bearer-token / JWT authentication bypass
     bugs in Node.js / Express middleware where invalid tokens leak
     protected user data. Use when an Express auth middleware
     (tokenValidator, roleChecker, tokenPolicy) defers errors,
     when /api/users-style routes return user records under bad
     tokens, when soft-fail / monitor-only auth modes mask
     401-worthy failures, or when the verifier reports user-data
     fields (email, phone, ssn_last4) leaking through 200
     responses."

   POOR example (does NOT trigger reliably):
     "Use when debugging a software bug. Provides a structured,
     step-by-step workflow for identifying the root cause."
     -- no domain keywords; "debug a software bug" matches
     everything = matches nothing.

3. **Recommended body sections** (omit ones you don't need):
   `# <title>` -> `## When to use` -> `## Workflow` -> `## Examples`
   -> `## Gotchas` -> `## Output template`.
   Gotchas is usually the single highest-value section -- write it
   when the trajectory shows any near-miss or recovery.

4. **Pick ONE default path** in the workflow; relegate alternatives to
   a brief "Alternatives" line.

5. **Tier-3 files** are optional. Add a `<slug>/scripts/<file>` only
   when the trajectory shows the agent re-deriving the same logic or
   the step is fragile enough that a tested script beats freeform
   generation. Use `<slug>/references/<file>` for long material loaded
   on demand. Use `<slug>/assets/<file>` for verbatim templates. Every
   Tier-3 file MUST be cited from SKILL.md body with a clear trigger;
   uncited Tier-3 files are dead weight.

## Required JSON output

```
{
  "summary": "<one short sentence>",
  "operation_type": "create",
  "upsert_files": {
    "<slug>/SKILL.md": "<full SKILL.md, YAML frontmatter first>",
    "<slug>/scripts/<file>": "<optional, only if SKILL.md cites it>",
    "<slug>/references/<file>": "<optional, only if SKILL.md cites it>",
    "<slug>/assets/<file>": "<optional, only if SKILL.md cites it>"
  }
}
```

Minimal acceptable output is just the `<slug>/SKILL.md` entry.
"""


_INDUCE_USER_TEMPLATE = """\
## Family id

{family_id}

## Latent skill id (formal, internal handle)

{latent_skill_id}

## Slug (folder name AND frontmatter `name` -- MUST be identical)

{slug}

## Last task verdict

verifier_passed: {verifier_passed}
failure_summary: {failure_summary}

## Compacted trajectory (your only ground-truth evidence)

```
{trajectory_summary}
```
"""
\end{lstlisting}

\noindent\textbf{Zero-shot generation prompt templates}\par\smallskip
\begin{lstlisting}[style=sevbpromptstyle]
_ZERO_SHOT_SYSTEM = """\
You are a skill-evolution assistant maintaining a shared library of
procedural skills used by an LLM agent. Your role on this call is to
DRAFT a procedural skill BEFORE the agent has executed any task in
this family -- you only have the family label and a brief description,
NO execution trace yet. Output STRICT JSON.

(Mindset note for ZERO-SHOT: cast a WIDE net on triggers and workflow
shape, but do NOT invent project-specific gotchas / examples / numbers
-- you have no evidence yet. T1's induction will fill those in once a
real trace exists. The "comprehensive + adversarial" mindset below
applies to the GENERAL category, not to invented specifics.)

""" + _SKILL_MINDSET + _SPEC_BLOCK + """

================================================================================
HOW TO PROCESS A ZERO-SHOT DRAFT REQUEST
================================================================================

The user message provides:
  ## Family id, latent skill id, slug
  ## Family name + description (your only inputs -- no trace, no outcome)

Apply these zero-shot rules (in addition to the spec block above):

You are writing Tier 1 + Tier 2 ONLY. Tier 3 (`scripts/`, `references/`,
`assets/`) MUST NOT be created yet -- wait until T1's real execution
reveals what's worth caching.

1. **No invented gotchas, no invented examples.** You have no execution
   evidence. Inventing project quirks would mislead the agent. Leave
   `## Gotchas` and `## Examples` for T1 induction to fill in from the
   real trace.

2. **Frontmatter** -- `name` and `description` (HARD constraints):

   `name` MUST equal the slug given in the user message.

   `description` is the entire trigger surface (Tier-1, always in
   future agents' context). REQUIREMENTS:

   a. <=1024 chars, imperative voice ("Use this skill when...").
   b. **Keyword density**: extract AT LEAST 5 concrete keywords
      from the FAMILY description / family name provided in the
      user message -- domain terms, tool names, error patterns,
      file types, library names. Generic words ("debug", "fix",
      "code", "bug") DO NOT count.
   c. **State WHAT (capability) AND WHEN (trigger contexts)**.
      Cast WIDE here (T1 induction will narrow once a real trace
      exists). List 3-5 trigger contexts including INDIRECT ones
      where the user does NOT name the skill domain.
   d. NO meta-talk about the skill itself ("a structured workflow
      for...", "this skill helps with...") -- dilutes signal.

   POOR example (does NOT trigger reliably):
     "Use when debugging a software bug. Provides a structured,
     step-by-step workflow for identifying the root cause."
     -- generic, no domain keywords; matches everything = matches
     nothing.

3. **Body** (only the sections you can honestly write without a trace):
   - `# <title>` -- one line.
   - `## When to use` -- bullet list of trigger conditions (cast wider
     than the description).
   - `## Workflow` -- best-guess numbered procedure based on the family
     description. Prescriptive on obviously fragile / order-sensitive
     steps; permissive on creative ones.
   - `## Output template` -- only if the family description implies a
     fixed output shape; otherwise omit.

4. **One default path, no menus.** If multiple approaches are plausible,
   pick the most common one and move on.

5. **Keep the body under ~200 lines.** Post-T1 revision will refine it;
   over-investing now will mostly be rewritten.

## Required JSON output

```
{
  "summary": "<one short sentence>",
  "operation_type": "create",
  "upsert_files": {
    "<slug>/SKILL.md": "<full SKILL.md content, YAML frontmatter first>"
  }
}
```

Legacy single-key form is also accepted for backward compatibility:
`{"summary": "...", "skill_md": "<SKILL.md content>"}`.
"""


_ZERO_SHOT_USER_TEMPLATE = """\
## Family id

{family_id}

## Latent skill id (formal, internal handle)

{latent_skill_id}

## Slug (folder name AND frontmatter `name` -- MUST be identical)

{slug}

## Family name

{name}

## Family description

{description}

## Output format

Return your response as a single JSON object exactly per the schema in the system prompt.
"""
\end{lstlisting}

\subsection{Tier-3 Resource-Bundling Prompt and Parser Constraint}
\label{app:tier3_prompt_comparison}

The Tier-3 capacity ablation uses the same \texttt{Skill Author} call surface as the
main revision setting, but changes the revision instruction and parser
constraint. The default \texttt{free\_form} strategy allows the Skill Author
to add \texttt{scripts/}, \texttt{references/}, or \texttt{assets/} files when
they materially help. By contrast, the \texttt{tier3\_required} strategy
forces each eligible revision to include at least one such resource under an
existing skill. This creates a controlled comparison between ordinary
free-form skill editing and aggressive resource bundling, while keeping the
trajectory evidence, verifier feedback, existing skills, and Skill Author call
surface unchanged.

\newtcblisting{skillpromptbox}[2][]{
  enhanced,
  breakable,
  listing only,
  colback=gray!3,
  colframe=black!50,
  coltitle=white,
  colbacktitle=black!72,
  title={#2},
  fonttitle=\bfseries\small,
  boxed title style={
    sharp corners,
    boxrule=0pt,
  },
  arc=1mm,
  boxrule=0.45pt,
  left=1.4mm,
  right=1.4mm,
  top=1.0mm,
  bottom=1.0mm,
  listing options={
    basicstyle=\ttfamily\footnotesize,
    breaklines=true,
    columns=fullflexible,
    keepspaces=true,
    showstringspaces=false,
    upquote=true
  },
  #1
}

\begin{skillpromptbox}{Shared structured edit interface}
Return EXACTLY one JSON object with these keys:

{
  "summary": "<one short sentence describing the change>",
  "operation_type": "revise" | "narrow" | "replace" | "create",
  "upsert_files": {
    "<slug>/SKILL.md": "<full new SKILL.md content>",
    "<slug>/scripts/<file>": "<optional, only if SKILL.md cites it>",
    "<slug>/references/<file>": "<optional, only if SKILL.md cites it>",
    "<slug>/assets/<file>": "<optional, only if SKILL.md cites it>"
  }
}

Always include FULL file content -- never diffs. Include only files you
actually edited or created. No commentary outside the JSON.
\end{skillpromptbox}

\begin{skillpromptbox}{Without forced Tier-3: free-form revision mode}
Make whatever edit best closes the gap exposed by the failure.
Edit SKILL.md and -- when it materially helps -- add executable
`scripts/`, on-demand `references/`, or `assets/` files.
Cite every Tier-3 file you add from the SKILL.md body with a
clear "when to read / run" trigger; uncited files will be rejected.
Do NOT inflate the body for its own sake; size the change to the gap.
\end{skillpromptbox}

\begin{skillpromptbox}{With forced Tier-3: tier3-required revision mode}
MANDATORY for this revision: upsert_files MUST include at least ONE new
or updated file under `<slug>/scripts/`, `<slug>/references/`, or
`<slug>/assets/`. A SKILL.md-only patch will be REJECTED by the parser.

For folder choice + naming + hard rules, follow system spec section 6.
The summary below is just a decision guide for picking the folder;
spec 6 is the binding contract for HOW to write the file.

DECISION GUIDE for this trial:
  - Trace shows the agent re-deriving fragile multi-step logic, OR running
    the same ad-hoc command repeatedly
      -> bundle as `<slug>/scripts/<verb_object>.py`
         and cite the exact command from SKILL.md.
  - Trace shows the agent looking up long enum / API error table / vendored
    doc that would bloat SKILL.md
      -> move to `<slug>/references/<topic>.md` and cite the trigger
         ("read this if X happens") in SKILL.md.
  - Trace shows the agent reproducing a fixed output structure
    (template / schema / config skeleton)
      -> bundle as `<slug>/assets/<name>.<ext>` and tell SKILL.md to copy/fill.

If a suitable Tier-3 file already exists in the target skill, UPDATE it in
place -- do NOT create a near-duplicate.

ADDITIVE BIAS still applies: cite this Tier-3 file from SKILL.md with a clear
"when to run / read / copy" trigger. Uncited files are dead weight and rejected.
\end{skillpromptbox}

The parser enforces this distinction. In \texttt{tier3\_required} mode,
\texttt{upsert\_files} must contain at least one Tier-3 path under an existing
target skill. A \texttt{SKILL.md}-only revision is invalid for this setting.
Thus, the capacity diagnostic tests whether forcing richer skill artifacts
improves reuse beyond ordinary free-form revision, while holding the evidence
source and authoring interface fixed.
\section{Full Tier-3 Capacity Ablation Results}
\label{app:tier3_full_results}

Finally, this section reports the full per-model and per-metric results for the Tier-3 capacity diagnostic in \Cref{sec:capacity_diagnostic}. 
The main text summarizes the central pattern: forced Tier-3 resource bundling increases library size, but does not reliably improve frozen deployment success. 
Here we provide the complete ablation table for the always-update conditions, including acquisition success, replay success, frozen deployment success, and the three deployment-role metrics.

\begin{table}[h]
\centering
\caption{
\textbf{\textsc{No-Skill} vs. always-update Tier-3 ablations.}
Values are success rates (\%).
Deltas report percentage-point differences from the corresponding \textsc{No-Skill} result.
For RSR, the reference is \textsc{No-Skill} LSR because \textsc{No-Skill} has no replay phase.
Red/blue/gray denote positive/negative/zero deltas.
}
\label{tab:noskill_vs_always_tier3_compact}
\scriptsize
\setlength{\tabcolsep}{1.35pt}
\renewcommand{\arraystretch}{1.08}
\begin{adjustbox}{max width=\linewidth}
\begin{tabular}{@{}cclrrrrrrrrrrrr@{}}
\toprule
\textbf{Agent Harness}
& \textbf{Model}
& \textbf{Condition}
& \multicolumn{2}{c}{\textbf{LSR}}
& \multicolumn{2}{c}{\textbf{RSR}}
& \multicolumn{2}{c}{\textbf{ESR}}
& \multicolumn{2}{c}{\textbf{CSSR}}
& \multicolumn{2}{c}{\textbf{ARSR}}
& \multicolumn{2}{c}{\textbf{CompSR}} \\
\cmidrule(lr){4-5}
\cmidrule(lr){6-7}
\cmidrule(lr){8-9}
\cmidrule(lr){10-11}
\cmidrule(lr){12-13}
\cmidrule(lr){14-15}
& &
& \textbf{\%} & \(\Delta\)
& \textbf{\%} & \(\Delta\)
& \textbf{\%} & \(\Delta\)
& \textbf{\%} & \(\Delta\)
& \textbf{\%} & \(\Delta\)
& \textbf{\%} & \(\Delta\) \\
\midrule
\harnesscell{20}{\textsc{Claude Code}} & \multirow{5}{*}{Claude Opus 4.6} & \cond{No-Skill} & 38.9 & \NA & -- & \NA & 37.8 & \NA & 46.7 & \NA & 36.7 & \NA & 30.0 & \NA \\
 &  & \cond{Curated Always} & 42.2 & \pos{3.3} & 46.7 & \pos{7.8} & 37.8 & \zero{0.0} & 40.0 & \down{-6.7} & 46.7 & \pos{10.0} & 26.7 & \down{-3.3} \\
 &  & \cond{Curated Always+Tier3} & 45.6 & \pos{6.7} & 42.2 & \pos{3.3} & 35.6 & \down{-2.2} & 43.3 & \down{-3.3} & 40.0 & \pos{3.3} & 23.3 & \down{-6.7} \\
 &  & \cond{SelfGen Always} & 42.2 & \pos{3.3} & 45.6 & \pos{6.7} & 37.8 & \zero{0.0} & 43.3 & \down{-3.3} & 46.7 & \pos{10.0} & 23.3 & \down{-6.7} \\
 &  & \cond{SelfGen Always+Tier3} & 44.4 & \pos{5.6} & 51.1 & \pos{12.2} & 40.0 & \pos{2.2} & 46.7 & \zero{0.0} & 43.3 & \pos{6.7} & 30.0 & \zero{0.0} \\
\addlinespace[0.10em]
\cdashline{2-15}
\addlinespace[0.10em]
 & \multirow{5}{*}{Claude Opus 4.5} & \cond{No-Skill} & 42.2 & \NA & -- & \NA & 32.2 & \NA & 40.0 & \NA & 40.0 & \NA & 16.7 & \NA \\
 &  & \cond{Curated Always} & 42.2 & \zero{0.0} & 44.4 & \pos{2.2} & 36.7 & \pos{4.4} & 43.3 & \pos{3.3} & 43.3 & \pos{3.3} & 23.3 & \pos{6.7} \\
 &  & \cond{Curated Always+Tier3} & 45.6 & \pos{3.3} & 45.6 & \pos{3.3} & 32.2 & \zero{0.0} & 33.3 & \down{-6.7} & 36.7 & \down{-3.3} & 26.7 & \pos{10.0} \\
 &  & \cond{SelfGen Always} & 41.1 & \down{-1.1} & 40.0 & \down{-2.2} & 35.6 & \pos{3.3} & 40.0 & \zero{0.0} & 40.0 & \zero{0.0} & 26.7 & \pos{10.0} \\
 &  & \cond{SelfGen Always+Tier3} & 38.9 & \down{-3.3} & 38.9 & \down{-3.3} & 34.4 & \pos{2.2} & 46.7 & \pos{6.7} & 40.0 & \zero{0.0} & 16.7 & \zero{0.0} \\
\addlinespace[0.10em]
\cdashline{2-15}
\addlinespace[0.10em]
 & \multirow{5}{*}{Claude Sonnet 4.6} & \cond{No-Skill} & 37.8 & \NA & -- & \NA & 38.9 & \NA & 40.0 & \NA & 50.0 & \NA & 26.7 & \NA \\
 &  & \cond{Curated Always} & 40.0 & \pos{2.2} & 44.4 & \pos{6.7} & 38.9 & \zero{0.0} & 46.7 & \pos{6.7} & 43.3 & \down{-6.7} & 26.7 & \zero{0.0} \\
 &  & \cond{Curated Always+Tier3} & 41.1 & \pos{3.3} & 44.4 & \pos{6.7} & 36.7 & \down{-2.2} & 40.0 & \zero{0.0} & 43.3 & \down{-6.7} & 26.7 & \zero{0.0} \\
 &  & \cond{SelfGen Always} & 41.1 & \pos{3.3} & 46.7 & \pos{8.9} & 40.0 & \pos{1.1} & 40.0 & \zero{0.0} & 50.0 & \zero{0.0} & 30.0 & \pos{3.3} \\
 &  & \cond{SelfGen Always+Tier3} & 42.2 & \pos{4.4} & 44.4 & \pos{6.7} & 40.0 & \pos{1.1} & 43.3 & \pos{3.3} & 46.7 & \down{-3.3} & 30.0 & \pos{3.3} \\
\addlinespace[0.10em]
\cdashline{2-15}
\addlinespace[0.10em]
 & \multirow{5}{*}{Claude Sonnet 4.5} & \cond{No-Skill} & 41.1 & \NA & -- & \NA & 35.6 & \NA & 40.0 & \NA & 46.7 & \NA & 20.0 & \NA \\
 &  & \cond{Curated Always} & 37.8 & \down{-3.3} & 38.9 & \down{-2.2} & 33.3 & \down{-2.2} & 40.0 & \zero{0.0} & 40.0 & \down{-6.7} & 20.0 & \zero{0.0} \\
 &  & \cond{Curated Always+Tier3} & 40.0 & \down{-1.1} & 44.4 & \pos{3.3} & 35.6 & \zero{0.0} & 43.3 & \pos{3.3} & 40.0 & \down{-6.7} & 23.3 & \pos{3.3} \\
 &  & \cond{SelfGen Always} & 38.9 & \down{-2.2} & 44.4 & \pos{3.3} & 33.3 & \down{-2.2} & 40.0 & \zero{0.0} & 40.0 & \down{-6.7} & 20.0 & \zero{0.0} \\
 &  & \cond{SelfGen Always+Tier3} & 40.0 & \down{-1.1} & 36.7 & \down{-4.4} & 30.0 & \down{-5.6} & 33.3 & \down{-6.7} & 33.3 & \down{-13.3} & 23.3 & \pos{3.3} \\
\midrule
\harnesscell{15}{\textsc{Codex CLI}} & \multirow{5}{*}{GPT-5.4} & \cond{No-Skill} & 43.3 & \NA & -- & \NA & 33.3 & \NA & 43.3 & \NA & 33.3 & \NA & 23.3 & \NA \\
 &  & \cond{Curated Always} & 43.3 & \zero{0.0} & 42.2 & \down{-1.1} & 40.0 & \pos{6.7} & 40.0 & \down{-3.3} & 50.0 & \pos{16.7} & 30.0 & \pos{6.7} \\
 &  & \cond{Curated Always+Tier3} & 44.4 & \pos{1.1} & 40.0 & \down{-3.3} & 34.4 & \pos{1.1} & 46.7 & \pos{3.3} & 33.3 & \zero{0.0} & 23.3 & \zero{0.0} \\
 &  & \cond{SelfGen Always} & 44.4 & \pos{1.1} & 43.3 & \zero{0.0} & 35.6 & \pos{2.2} & 46.7 & \pos{3.3} & 40.0 & \pos{6.7} & 20.0 & \down{-3.3} \\
 &  & \cond{SelfGen Always+Tier3} & 44.4 & \pos{1.1} & 46.7 & \pos{3.3} & 32.2 & \down{-1.1} & 40.0 & \down{-3.3} & 36.7 & \pos{3.3} & 20.0 & \down{-3.3} \\
\addlinespace[0.10em]
\cdashline{2-15}
\addlinespace[0.10em]
 & \multirow{5}{*}{GPT-5.3-Codex} & \cond{No-Skill} & 44.4 & \NA & -- & \NA & 34.4 & \NA & 36.7 & \NA & 46.7 & \NA & 20.0 & \NA \\
 &  & \cond{Curated Always} & 42.2 & \down{-2.2} & 45.6 & \pos{1.1} & 32.2 & \down{-2.2} & 40.0 & \pos{3.3} & 40.0 & \down{-6.7} & 16.7 & \down{-3.3} \\
 &  & \cond{Curated Always+Tier3} & 42.2 & \down{-2.2} & 47.8 & \pos{3.3} & 37.8 & \pos{3.3} & 46.7 & \pos{10.0} & 43.3 & \down{-3.3} & 23.3 & \pos{3.3} \\
 &  & \cond{SelfGen Always} & 44.4 & \zero{0.0} & 48.9 & \pos{4.4} & 34.4 & \zero{0.0} & 36.7 & \zero{0.0} & 43.3 & \down{-3.3} & 23.3 & \pos{3.3} \\
 &  & \cond{SelfGen Always+Tier3} & 46.7 & \pos{2.2} & 50.0 & \pos{5.6} & 33.3 & \down{-1.1} & 36.7 & \zero{0.0} & 43.3 & \down{-3.3} & 20.0 & \zero{0.0} \\
\addlinespace[0.10em]
\cdashline{2-15}
\addlinespace[0.10em]
 & \multirow{5}{*}{GPT-5.2-Codex} & \cond{No-Skill} & 43.3 & \NA & -- & \NA & 36.7 & \NA & 40.0 & \NA & 53.3 & \NA & 16.7 & \NA \\
 &  & \cond{Curated Always} & 44.4 & \pos{1.1} & 45.6 & \pos{2.2} & 37.8 & \pos{1.1} & 46.7 & \pos{6.7} & 43.3 & \down{-10.0} & 23.3 & \pos{6.7} \\
 &  & \cond{Curated Always+Tier3} & 46.7 & \pos{3.3} & 45.6 & \pos{2.2} & 35.6 & \down{-1.1} & 46.7 & \pos{6.7} & 43.3 & \down{-10.0} & 16.7 & \zero{0.0} \\
 &  & \cond{SelfGen Always} & 46.7 & \pos{3.3} & 46.7 & \pos{3.3} & 38.9 & \pos{2.2} & 50.0 & \pos{10.0} & 43.3 & \down{-10.0} & 23.3 & \pos{6.7} \\
 &  & \cond{SelfGen Always+Tier3} & 46.7 & \pos{3.3} & 44.4 & \pos{1.1} & 34.4 & \down{-2.2} & 40.0 & \zero{0.0} & 40.0 & \down{-13.3} & 23.3 & \pos{6.7} \\
\midrule
\harnesscell{15}{\textsc{Gemini CLI}} & \multirow{5}{*}{Gemini 3.1 Pro} & \cond{No-Skill} & 40.0 & \NA & -- & \NA & 35.6 & \NA & 40.0 & \NA & 46.7 & \NA & 20.0 & \NA \\
 &  & \cond{Curated Always} & 40.0 & \zero{0.0} & 45.6 & \pos{5.6} & 35.6 & \zero{0.0} & 40.0 & \zero{0.0} & 40.0 & \down{-6.7} & 26.7 & \pos{6.7} \\
 &  & \cond{Curated Always+Tier3} & 45.6 & \pos{5.6} & 41.1 & \pos{1.1} & 40.0 & \pos{4.4} & 43.3 & \pos{3.3} & 50.0 & \pos{3.3} & 26.7 & \pos{6.7} \\
 &  & \cond{SelfGen Always} & 42.2 & \pos{2.2} & 41.1 & \pos{1.1} & 33.3 & \down{-2.2} & 33.3 & \down{-6.7} & 43.3 & \down{-3.3} & 23.3 & \pos{3.3} \\
 &  & \cond{SelfGen Always+Tier3} & 45.6 & \pos{5.6} & 45.6 & \pos{5.6} & 35.6 & \zero{0.0} & 40.0 & \zero{0.0} & 40.0 & \down{-6.7} & 26.7 & \pos{6.7} \\
\addlinespace[0.10em]
\cdashline{2-15}
\addlinespace[0.10em]
 & \multirow{5}{*}{Gemini 3 Flash} & \cond{No-Skill} & 40.0 & \NA & -- & \NA & 35.6 & \NA & 30.0 & \NA & 53.3 & \NA & 23.3 & \NA \\
 &  & \cond{Curated Always} & 43.3 & \pos{3.3} & 37.8 & \down{-2.2} & 37.8 & \pos{2.2} & 46.7 & \pos{16.7} & 36.7 & \down{-16.7} & 30.0 & \pos{6.7} \\
 &  & \cond{Curated Always+Tier3} & 38.9 & \down{-1.1} & 41.1 & \pos{1.1} & 33.3 & \down{-2.2} & 36.7 & \pos{6.7} & 43.3 & \down{-10.0} & 20.0 & \down{-3.3} \\
 &  & \cond{SelfGen Always} & 35.6 & \down{-4.4} & 40.0 & \zero{0.0} & 36.7 & \pos{1.1} & 36.7 & \pos{6.7} & 43.3 & \down{-10.0} & 30.0 & \pos{6.7} \\
 &  & \cond{SelfGen Always+Tier3} & 43.3 & \pos{3.3} & 40.0 & \zero{0.0} & 27.8 & \down{-7.8} & 33.3 & \pos{3.3} & 36.7 & \down{-16.7} & 13.3 & \down{-10.0} \\
\addlinespace[0.10em]
\cdashline{2-15}
\addlinespace[0.10em]
 & \multirow{5}{*}{Gemini 2.5 Pro} & \cond{No-Skill} & 30.0 & \NA & -- & \NA & 26.7 & \NA & 26.7 & \NA & 30.0 & \NA & 23.3 & \NA \\
 &  & \cond{Curated Always} & 31.1 & \pos{1.1} & 26.7 & \down{-3.3} & 24.4 & \down{-2.2} & 26.7 & \zero{0.0} & 23.3 & \down{-6.7} & 23.3 & \zero{0.0} \\
 &  & \cond{Curated Always+Tier3} & 30.0 & \zero{0.0} & 27.8 & \down{-2.2} & 25.6 & \down{-1.1} & 33.3 & \pos{6.7} & 23.3 & \down{-6.7} & 20.0 & \down{-3.3} \\
 &  & \cond{SelfGen Always} & 31.1 & \pos{1.1} & 34.4 & \pos{4.4} & 25.6 & \down{-1.1} & 40.0 & \pos{13.3} & 20.0 & \down{-10.0} & 16.7 & \down{-6.7} \\
 &  & \cond{SelfGen Always+Tier3} & 23.3 & \down{-6.7} & 21.1 & \down{-8.9} & 27.8 & \pos{1.1} & 36.7 & \pos{10.0} & 26.7 & \down{-3.3} & 20.0 & \down{-3.3} \\
\bottomrule
\end{tabular}
\end{adjustbox}
\end{table}

\Cref{tab:noskill_vs_always_tier3_compact} compares \textsc{Curated-Always} and \textsc{SelfGen-Always} with their Tier-3 variants against the corresponding \textsc{No-Skill} baselines. 
The table shows that Tier-3 resource bundling can improve some local metrics, especially LSR or RSR, but those gains do not consistently carry over to ESR, CSSR, ARSR, or CompSR. 
For example, some models benefit from \textsc{SelfGen-Always+Tier3} or \textsc{Curated-Always+Tier3}, while others lose frozen deployment performance despite larger libraries. 
This supports the main-text conclusion that the limiting factor is not simply storage capacity, but selective resource persistence: agents must decide which trace-derived details are stable enough to become reusable procedural resources.

\end{document}